\pdfoutput=1

\documentclass[11pt]{article}
\usepackage[final]{emnlp2022}
\usepackage{times}
\usepackage{latexsym}
\usepackage[T1]{fontenc}

\usepackage[utf8]{inputenc}
\usepackage{graphicx}
\usepackage{multirow}
\usepackage{float}
\usepackage{cleveref}

\usepackage{microtype}
\everypar{\looseness=-1}
\title{A Federated Approach to Predicting Emojis in Hindi Tweets}

\author{Deep Gandhi*$^1$, Jash Mehta*$^2$, Nirali Parekh$^3$, Karan Waghela$^4$,\\ {\bf Lynette D'Mello$^5$}, {\bf Zeerak Talat$^6$}\\
$^1$University of Alberta, $^2$Georgia Institute of Technology
$^3$Stanford University\\
$^4$Santa Clara University, $^5$DJ Sanghvi College of Engineering, $^6$Simon Fraser University\\
\texttt{$^1$drgandhi@ualberta.ca, $^2$jmehta73@gatech.edu, $^3$nirali25@stanford.edu}\\
\texttt{$^4$kwaghela@scu.edu, $^5$lynette.dmello@djsce.ac.in, $^6$zeerak\_talat@sfu.ca}}

\begin{document}
\maketitle
\def\thefootnote{*}\footnotetext{Equal contribution.}\def\thefootnote{\arabic{footnote}}
\begin{abstract}
The use of emojis affords a visual modality to, often private, textual communication.
The task of predicting emojis however provides a challenge for machine learning as emoji use tends to cluster into the frequently used and the rarely used emojis.
Much of the machine learning research on emoji use has focused on high resource languages and has conceptualised the task of predicting emojis around traditional server-side machine learning approaches.
However, traditional machine learning approaches for private communication can introduce privacy concerns, as these approaches require all data to be transmitted to a central storage.
In this paper, we seek to address the dual concerns of emphasising high resource languages for emoji prediction and risking the privacy of people's data.
We introduce a new dataset of $118$k tweets (augmented from $25$k unique tweets) for emoji prediction in Hindi,\footnote{The dataset and code can be accessed at https://github.com/deep1401/fedmoji} and propose a modification to the federated learning algorithm, CausalFedGSD, which aims to strike a balance between model performance and user privacy. We show that our approach obtains comparative scores with more complex centralised models while reducing the amount of data required to optimise the models and minimising risks to user privacy.
\end{abstract}

\section{Introduction}
Since the creation of emojis around the turn of the millennium \citep{stark2015conservatism,alshenqeeti2016emojis}, they have become of a staple of informal textual communication, expressing emotion and intent in written text \citep{barbieri-etal-2018-interpretable}.
This development in communication style has prompted research into emoji analysis and prediction for English \citep[e.g.][]{barbieri-etal-2018-semeval,barbieri-etal-2018-interpretable,felbo-etal-2017-using,tomihira2020multilingual,zhang2020emoji}.
Comparatively little research attention has been given to the low resource languages.

\begin{figure}
\centering
\includegraphics[scale=0.35]{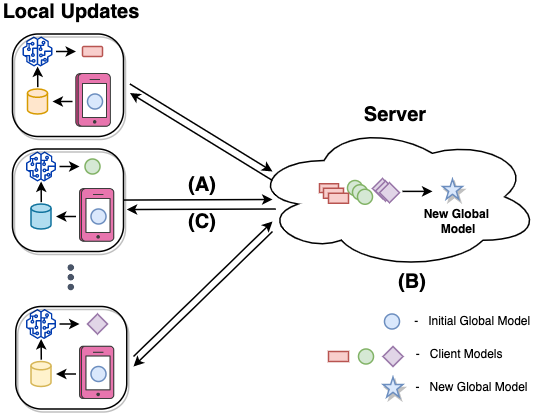}
\caption{The Federated Learning process: (A) client devices compute weight updates on locally stored data, (B) client weight updates are transmitted to the server and used to update the global model, (C) the resulting global model is redistributed to all clients.} \label{fig1}
\end{figure}

Emoji-prediction has posed a challenge for the research community because emojis express multiple modalities, contain visual semantics and the ability to stand in place for words~\citep{padilla-lopez-cap-2017-ever}.
The challenge is further compounded by the quantity of emojis sent and the imbalanced distribution of emoji use~\citep{cappallo2018new,padilla-lopez-cap-2017-ever}.
Machine learning (ML) for emoji analysis and prediction has traditionally relied on traditional server-side architectures. However, training such models risks leaking sensitive information that may co-occur with emojis or be expressed through them.
This can lead to potential breaches of data privacy regulation (e.g. the European General Data Protection Regulation and the California Consumer Privacy Act).
In contrast, federated learning (FL)~\citep{mcmahan2017communication} approaches the task of training machine learning models by emphasising privacy of data.
Such privacy is ensured by training models locally and sharing weight updates, rather than the data, with a central server (see Figure~\ref{fig1}).
The FL approach assumes that some client-updates may be corrupted during transmission. 
FL therefore aims to retain predictive performance while emphasising user privacy in scenarios with potential data loss.

Motivated by prior work in privacy preserving ML~\citep[e.g.][]{ramaswamy2019federated,yang2018applied} and emoji prediction for low resource languages~\citep[e.g.][]{choudhary2018contrastive}, we examine the application of FL to emoji topic prediction for Hindi.
Specifically, we collect an imbalanced dataset of $118,030$ tweets in Hindi which contain $700$ unique emojis that we classify into $10$ pre-defined categories of emojis.
The dataset contains $700$ unique emojis, that we classify into $10$ pre-defined categories of emojis.\footnote{These categories are obtained from the Emojis library, available at https://github.com/alexandrevicenzi/emojis.}
We further examine the impact of two different data balancing strategies on federated and server-side, centralised model performance.
Specifically, we examine: re-sampling and cost-sensitive re-weighting.
We consider $6$ centralised models which form our baselines:
Bi-directional LSTM~\citep{Hochreiter_Schmidhuber_1997}, IndicBert~\citep{kakwani-etal-2020-indicnlpsuite}, HindiBERT,\footnote{https://huggingface.co/monsoon-nlp/hindi-bert} Hindi-Electra,\footnote{https://huggingface.co/monsoon-nlp/hindi-tpu-electra} mBERT~\citep{devlin-etal-2019-bert}, and XLM-R~\citep{conneau-etal-2020-unsupervised}; and LSTMs trained using two FL algorithms: FedProx~\citep{li2018federated} and a modified version of CausalFedGSD~\citep{francis2021towards}.

We show that LSTMs trained using FL perform competitively with more complex, centralised models in spite of only using up to $50\%$ of the data.

\section{Prior work}
\paragraph{Federated Learning}
Federated Learning~\citep[FL,][]{mcmahan2017communication} is a training procedure that distributes training of models onto a number of client devices.
Each client device locally computes weight updates on the basis of local data, and transmits the updated weights to a central server.
In this way, FL can help prevent computational bottlenecks when training models on a large corpus while simultaneously preserving privacy by not transmitting raw data.
This training approach has previously been applied for on-device token prediction on mobile phones for English.
In a study of the quality of mobile keyboard suggestions, \citet{yang2018applied} show that FL improves the quality of suggested words.
Addressing emoji-prediction in English, \citet{ramaswamy2019federated} use the FederatedAveraging algorithm, to improve on traditional server-based models on user devices.
We diverge from \citet{ramaswamy2019federated} by using the CausalFedGSD and FedProx algorithms on Hindi tweets.
FedProx develops on the FederatedAveraging algorithm by introducing a regularization constant to it \cite{li2018federated}.
In related work, \citet{choudhary2018contrastive} seek to address the question of FL for emoji prediction for low resource languages.
However, the dataset that they use, \citet{choudhary-etal-2018-twitter} relies on emojis that are frequently used in English text and therefore may not be representative of emoji use in other, low resource languages.

\paragraph{Centralised Training}
In efforts to extend emoji prediction, \citet{ma2020emoji} experiment with a BERT-based model on a new English dataset that includes a large set of emojis for multi label prediction. 
Addressing the issue of low resource languages, \citet{choudhary2018contrastive} train a bi-directional LSTM-based siamese network, jointly training their model with high resource and low resource languages.
A number of studies on emoji prediction have been conducted in lower-resourced languages than English \citep[e.g.][]{chayahebrew,ronzano2018overview,choudhary-etal-2018-twitter,barbieri-etal-2018-semeval,duarte2020emoji,tomihira2020multilingual}.
Common to these approaches is the use of centralised ML models, which increase risks of breaches of privacy.
In our experiments, we study using FL for emoji topic classification in low resource settings.

\section{Data}
\label{sec:Dataset and Pre-processing}
\begin{figure}
\centering
\includegraphics[width=\columnwidth]{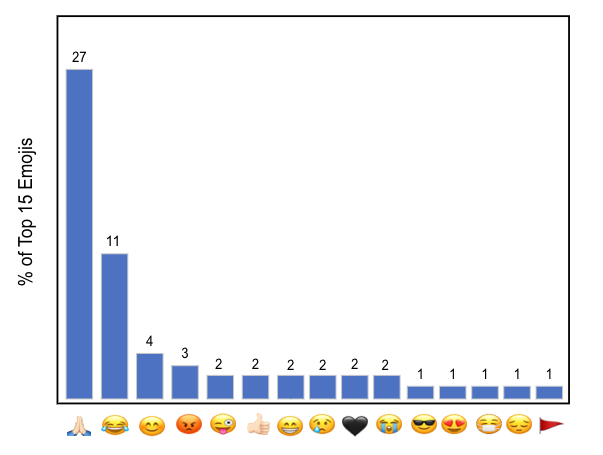}
\caption{Distribution of 15 most frequently appearing emojis in our dataset.} \label{fig2}
\end{figure}

We collect our dataset for emoji topic prediction by scraping ${\sim}1$M tweets. 
We only keep the $24,794$ tweets that are written in Hindi and contain at least one emoji.
We duplicate all tweets that contain multiple emojis by the number of emojis contained, assigning a single emoji to each copy, resulting in a dataset of $118,030$ tweets with $700$ unique emojis.
Due to the imbalanced distribution of emojis in our dataset (see Figure \ref{fig2}), we assign emojis into $10$ coarse-grained categories.
This reduction i.e., from multi-label to multi-class and unique emojis into categories, risks losing the semantic meaning of emojis.
Our decision is motivated by how challenging emoji prediction is without such reductions \citep{choudhary2018contrastive}.\looseness=-1

We pre-process our data to limit the risk of over-fitting to rare tokens and platform specific tokens. For instance, we lower-case all text and remove numbers, punctuation, and retweet markers.
We replace mentions, URLs, and hashtags with specific tokens to avoid issues of over-fitting to these.

\subsection{Balancing data}\label{sec:balance}
This dataset exhibits a long-tail in the distribution of emoji categories (see Figure~\ref{fig3}), with the vast majority of tweets belonging to the ``Smileys \& Emotions'' and ``People \& Body'' categories.
To address this issue, we use two different data balancing methods: re-sampling~\citep{he2009learning} and cost-sensitive re-weighting~\citep{khan2017cost}.

\paragraph{Re-Sampling}
Re-sampling has been used widely to address issues of class imbalances~\citep[e.g.][]{buda2018systematic,zou2018unsupervised,geifman2017deep,shen2016relay}.
We balance the training data by up-sampling the minority class \citep{drumnond2003class} and down-sampling the majority class \citep{chawla2002smote}, resulting in a balanced dataset of $94,420$ tweets ($9,442$ documents per class).
The validation and test sets are left unmodified to ensure a fair and realistic evaluation.

\paragraph{Cost-Sensitive learning}
Another method for addressing data imbalances is cost-sensitive learning~\citep[see][]{zhou2005training,huang2016learning,ting2000comparative,sarafianos2018deep}.
In this method, each class is assigned a weight which is used to weigh the loss function \cite{lin2017focal}.
For our models, we set the class weights to the inverse class frequencies.

\section{Experiments}
We conduct experiments with PyTorch \citep{Paszke_et_al} and Transformers \citep{wolf-etal-2020-transformers} on Google Colab using a Nvidia Tesla V100 GPU with 26GB of RAM.
We create train, validation, and test ($80/10/10$) splits of the dataset, and measure performances using precision, recall, and weighted F1.
All models are trained and evaluated on the imbalanced data and the two data balancing methods (see \S\ref{sec:balance}).
For the FL setting, we conduct experiments manipulating the independent and identically distributed (I.I.D.) data assumption on client nodes.

\begin{table*}[t]
\resizebox{\textwidth}{!}{
\begin{tabular}{c|ccc|ccc|ccc|ccc|ccc|ccc}
& 
\multicolumn{3}{c|}{\textbf{Bi-LSTM}} & \multicolumn{3}{c|}{\textbf{mBERT}} &
\multicolumn{3}{c|}{\textbf{XLM-R}} &
\multicolumn{3}{c|}{\textbf{IndicBERT}} &
\multicolumn{3}{c|}{\textbf{hindiBERT}} &
\multicolumn{3}{c}{\textbf{Hindi-Electra}} \\
& Precision & Recall & F1
& Precision & Recall & F1          
& Precision & Recall & F1          
& Precision & Recall & F1             
& Precision & Recall & F1            
& Precision & Recall & F1\\ 
\hline
\textbf{Imbalanced}
& $64.72$ & $64.26$ & $63.83$
& $63.25$ & $66.90$ & $64.50$
& $\textbf{68.74}$ & $\textbf{70.39}$ & $\textbf{69.44}$
& $67.15$ & $68.22$ & $67.60$
& $65.39$ & $66.53$ & $65.90$
& $27.34$ & $52.29$ & $35.91$ \\ 
\textbf{Re-sampled}
& $64.42$ & $55.41$ & $58.61$         
& $62.18$ & $53.43$ & $56.58$
& $67.92$ & $60.76$ & $63.39$
& $68.04$ & $62.44$ & $64.58$ 
& $62.95$ & $55.16$ & $57.92$          
& $64.42$ & $57.93$ & $60.30$ \\ 
\textbf{Cost-Sensitive} 
& $68.41$ & $62.27$ & $64.46$              
& $63.99$ & $62.73$ & $63.30$ 
& $69.79$ & $68.33$ & $68.87$
& $69.54$ & $67.98$ & $68.66$
& $66.97$ & $65.32$ & $66.06$          
& $27.34$ & $52.29$ & $35.91$ \\ 
\end{tabular}
}
\caption{Centralised model performances.}
\label{server_results}
\end{table*}

\begin{table*}[]
\resizebox{\textwidth}{!}{
\begin{tabular}{c|ccc|ccc|ccc|ccc|ccc|ccc}

 & \multicolumn{6}{c|}{$\textbf{c = 10\%}$} 
 & \multicolumn{6}{c|}{$\textbf{c = 30\%}$} 
 & \multicolumn{6}{c}{$\textbf{c = 50\%}$} \\\hline
 & \multicolumn{3}{c|}{IID} & \multicolumn{3}{c|}{non-IID} 
 & \multicolumn{3}{c|}{IID} & \multicolumn{3}{c|}{non-IID} 
 & \multicolumn{3}{c|}{IID} & \multicolumn{3}{c}{non-IID} \\  
 & Precision    & Recall    & F1    & Precision      & Recall     & F1     & Precision    & Recall    & F1    & Precision      & Recall     & F1     & Precision    & Recall    & F1    & Precision      & Recall     & F1     \\ \hline
\textbf{Imbalanced} &
  $60.94$ &
  $66.11$ &
  $62.99$ &
  $61.05$ &
  $39.68$ &
  $25.82$ &
  $61.10$ &
  $66.01$ &
  $63.04$ &
  $57.88$ &
  $66.39$ &
  $61.64$ &
  $\textbf{61.11}$ &
  $\textbf{66.91}$ &
  $\textbf{63.35}$ &
  $56.94$ &
  $63.82$ &
  $57.06$ \\ 
\textbf{Re-sampled} &
  $60.89$ &
  $46.01$ &
  $50.89$ &
  $60.83$ &
  $22.52$ &
  $23.04$ &
  $60.58$ &
  $46.58$ &
  $51.22$ &
  $57.38$ &
  $35.85$ &
  $37.37$ &
  $60.78$ &
  $47.14$ &
  $51.63$ &
  $53.16$ &
  $36.83$ &
  $41.95$ \\ 
\textbf{Cost-Sensitive} &
  $60.45$ &
  $59.50$ &
  $59.50$ &
  $60.47$ &
  $41.24$ &
  $28.92$ &
  $60.91$ &
  $60.99$ &
  $60.47$ &
  $55.76$ &
  $56.77$ &
  $52.69$ &
  $61.56$ &
  $60.39$ &
  $60.48$ &
  $56.51$ &
  $63.41$ &
  $56.80$ \\ 
\end{tabular}
}
\caption{Results using the FedProx algorithm. c is the percentage of clients whose updates are considered.}
\label{fed_results}
\end{table*}

\begin{table*}[!h]
\resizebox{\textwidth}{!}{
\begin{tabular}{c|ccc|ccc|ccc|ccc|ccc|ccc}

 & \multicolumn{6}{c|}{$\textbf{c = 10\%}$} 
 & \multicolumn{6}{c|}{$\textbf{c = 30\%}$} 
 & \multicolumn{6}{c}{$\textbf{c = 50\%}$} \\\hline
 & \multicolumn{3}{c|}{IID} & \multicolumn{3}{c|}{non-IID} 
 & \multicolumn{3}{c|}{IID} & \multicolumn{3}{c|}{non-IID} 
 & \multicolumn{3}{c|}{IID} & \multicolumn{3}{c}{non-IID} \\  
 & Precision    & Recall    & F1    & Precision      & Recall     & F1     & Precision    & Recall    & F1    & Precision      & Recall     & F1     & Precision    & Recall    & F1    & Precision      & Recall     & F1     \\ \hline
\textbf{Imbalanced} &
  $60.89$ &
  $65.04$ &
  $62.34$ &
  $58.09$ &
  $60.84$ &
  $56.96$ &
  $61.40$ &
  $66.08$ &
  $\textbf{63.22}$ &
  $57.68$ &
  $\textbf{66.51}$ &
  $61.21$ &
  $60.69$ &
  $66.33$ &
  $62.90$ &
  $57.90$ &
  $57.84$ &
  $47.85$ \\ 
\textbf{Re-sampled} &
  $62.14$ &
  $44.95$ &
  $50.27$ &
  $53.31$ &
  $34.01$ &
  $39.69$ &
  $\textbf{62.46}$ &
  $45.39$ &
  $50.67$ &
  $58.21$ &
  $27.61$ &
  $27.61$ &
  $61.86$ &
  $46.00$ &
  $51.15$ &
  $59.84$ &
  $22.57$ &
  $28.83$ \\ 
\textbf{Cost-Sensitive} &
  $61.62$ &
  $61.15$ &
  $60.92$ &
  $56.72$ &
  $65.20$ &
  $60.51$ &
  $62.17$ &
  $61.97$ &
  $61.60$ &
  $56.86$ &
  $65.54$ &
  $60.59$ &
  $60.50$ &
  $61.28$ &
  $60.44$ &
  $59.33$ &
  $60.56$ &
  $54.22$ \\ 
\end{tabular}
}
\caption{Results using the modified CausalFedGSD. c is the percentage of clients whose updates are considered.}
\label{shared_fed_results}
\end{table*}

\begin{table}

\resizebox{\columnwidth}{!}{

\begin{tabular}{c|ccc|ccc|ccc}

& \multicolumn{3}{c|}{$\textbf{c = 10\%}$} 
& \multicolumn{3}{c|}{$\textbf{c = 30\%}$} 
& \multicolumn{3}{c}{$\textbf{c = 50\%}$} \\ 

& Precision & Recall & F1        
& Precision & Recall & F1             
& Precision & Recall & F1   \\ \hline

\textbf{Imbalanced}     
& $63.95$ & $64.19$ & $63.67$    
& $64.23$ & $64.44$ & $63.91$          
& $64.16$ & $64.28$ & $63.78$    \\ 

\textbf{Re-sampled}     
& $63.07$ & $51.14$ & $55.08$    
& $62.84$ & $52.04$ & $55.71$          
& $62.84$ & $51.72$ & $55.50$    \\ 

\textbf{Cost-Sensitive} 
& $66.72$ & $64.96$ & $65.38$    
& $66.66$ & $64.84$ & $65.27$ 
& $\textbf{66.78}$ & $\textbf{65.08}$ & $\textbf{65.47}$    \\ 

\end{tabular}
}

\caption{Results for the baseline CausalFedGSD. $c$ is the client fraction per round.}
\label{causalFedGSD}
\end{table}

\begin{table}
\resizebox{\columnwidth}{!}{
\begin{tabular}{lc|cc}
\hline \textbf{Approach} & Centralised & \multicolumn{2}{c} { Federated} \\
\hline & XLM-R & FedProx & Modified CausalFedGSD\\
\cline { 2 - 4 } 
Imbalanced & $69.44$ & $63.35$ & $63.22$\\
Re-sampled & $63.39$ & $51.63$ & $51.15$\\
Cost-Sensitive & $68.87$ & $60.48$ & $61.60$\\
\hline
\end{tabular}
}
\caption{An approach-wise comparison of F1 scores for best performing models in centralized and federated settings.}
\label{result_comparison}
\end{table}

\subsection{Baseline models}\label{sec:baseline_models}
We use $6$ centralised models as baselines for comparison with the federated approach.
Specifically, we use a bi-LSTM~\citep{Hochreiter_Schmidhuber_1997} with 2 hidden layers and dropout at $0.5$; two multi-lingual models: mBERT~\citep{devlin-etal-2019-bert} and XLM-R~\citep{conneau-etal-2020-unsupervised}; and three models pre-trained on Indic languages: IndicBert \citep{kakwani-etal-2020-indicnlpsuite}, HindiBERT, and Hindi-Electra.

\subsection{Federated models}
For our federated learning experiments, we use the FedProx~\citep{li2018federated} algorithm and a modification of the CausalFedGSD~\citep{francis2021towards} algorithm.
FedProx trains models by considering the dissimilarity between local gradients and adds a proximal term to the loss function to prevent divergence from non-I.I.D. data.
CausalfedGSD reserves $30\%$ of the training data for initializing client nodes.
When a client node is created, it receives a random sample of the reserved data for initialization.
In our modification, we similarly reserve $30\%$ of the training data, however we diverge by using the full sample to initialize the global model, which is then transmitted to client nodes.
This change means that (i) user data is transmitted fewer times; (ii) modellers retain control over the initialization of the model, e.g. to deal with class imbalances; and (iii) models converge faster, due to exposing client nodes to the distribution of all classes  (see Appendix \ref{sec:performance_comparison}).

We reuse the Bi-LSTM (see \Cref{sec:baseline_models}) as our experimental model on client devices due to its relative low compute requirements.
For our experiments, we set the number of clients to $100$ and simulate I.I.D. and non-I.I.D. settings.
We simulate an I.I.D. setting by ensuring that all client devices receive data that is representative of the entire dataset.
For the non-I.I.D. setting, we create severely imbalanced data splits for clients by first grouping the data by label, then splitting the grouped data into $200$ bins and randomly assigning $2$ bins to each client.
We experiment with three different settings, in which we randomly select $10\%$, $30\%$, and $50\%$ of all clients whose updates are incorporated into the global model.\looseness=-1

\subsection{Analysis}
Considering the results for our baseline models (see Table~\ref{server_results}), we find that XLM-R and IndicBERT obtain the best performances.
Further, using cost-sensitive weighting tends to out-perform re-sampling the dataset.
In fact, the cost-sensitive weighting performs comparatively, or out-performs, other settings.
Curiously, we see that Hindi Electra under-performs compared to all other models, including HindiBERT which is a smaller model trained on the same data.
This discrepancy in the performances models may be due to differences in complexity, and thus data required to achieve competitive performances.\footnote{The developers of Hindi Electra also note similar under-performance on other tasks.}
Finally, the bi-LSTM slightly under-performs in comparison to XLM-R, however it performs competitively with all other well-performing models.

Turning to the  performance of the federated baselines (see Table \ref{fed_results}), we find an expected performance of the models.\footnote{Please refer to the appendices for additional details on model performance and training.}
Generally, we find that the federated models achieve comparative performances, that are slightly lower than the centralised systems.
This is due to the privacy-performance trade-off, where the increased privacy offsets a small performance loss.
Considering the F1-scores, we find that the optimal setting of the ratio of clients is subject to the data being I.I.D.
In contrast, models trained on the re-sampled data tend to prefer data in an I.I.D. setting, but in general under-perform in comparison with other weighting strategies, including the imbalanced sample.
Using our modification of the CausalFedGSD algorithm, we show improvements over our FL baselines when the data is I.I.D. and variable performance for a non-I.I.D. setting (see Table~\ref{shared_fed_results}). Comparing the results of the best performing settings, we find that the FL architectures perform comparably with the centralised models, in spite of being exposed to less data and preserving privacy of users (see Table~\ref{result_comparison}).
Table~\ref{causalFedGSD} refers to the results for the I.I.D. experiments of the baseline CausalFedGSD algorithm~\citep{francis2021towards}. We also observe a difference in optimization time for both models (see Appendix \ref{sec:performance_comparison}). 
Models trained using our modification of CausalFedGSD converges faster than the original CausalFedGSD, which in turn converges much faster than FedProx.
Moreover, we find indications that the original CausalFedGSD algorithm may be prone to over-fitting, as performance stagnates without fluctuations, while our modification shows fluctuations in performance that are similar to those of the FedProx models.

\section{Conclusion}
Emoji topic prediction in user-generated text is a task which can contains highly private data.
It is therefore important to consider privacy-preserving methods for the task. 
Here, we presented a new dataset for the task for Hindi and compared a privacy preserving approach, Federated Learning, with the centralised server-trained method.
We present a modification to the CausalFedGSD algorithm, and find that it converges faster than our other experimetnal models.
In our experiments with different data balancing methods and simulations of I.I.D. and non-I.I.D. settings, we find that using FL affords comparable performances to the more complex fine-tuned language models that are trained centrally, while ensuring privacy.
In future work, we plan to extend this work to multi-label emoji topic prediction and investigate strategies for dealing with decay of the model vocabulary.

\section*{Ethical considerations}
The primary reason for using federated learning is to ensure user-privacy. 
The approach can then stand in conflict with open and reproducible science, in terms of data sharing.
We address this issue by making our dataset open to the public, given that researchers provide an Institutional Review Board (IRB) approval and a research statement that details the methods and goals of the research, where IRB processes are not implemented.
For researchers who are at institutions without IRB processes, data will only be released given a research statement that also details potential harms to participants.
The sharing of data will follow Twitter's developer agreement, which allows for $50$k Tweet objects to be shared. We will further provide the code to our $24$k tweets into the full dataset of $118$k.

Our modification of the CausalFedGSD model introduces the concern of some data being used to initialise the model.
Here a concern can be that some data will be available globally.
While this concern is justified, the use of FL affords two things:
First, FL can limit on the overall amount of raw data that is transmitted and risks exposure.
Second, initialisation can occur using synthetic data, created for the express purposes of model initialisation.
Moreover, pre-existing public, or privately owned, datasets can be used to initialise models, which can be further trained given weight updates provided by the client nodes.
Federated learning, and our approach to FL thus reduce the risks of exposing sensitive information about users, although the method does not completely remove such risks.

\bibliography{anthology,custom, appendix}

\begin{thebibliography}{40}
\expandafter\ifx\csname natexlab\endcsname\relax\def\natexlab#1{#1}\fi

\bibitem[{Alshenqeeti(2016)}]{alshenqeeti2016emojis}
Hamza Alshenqeeti. 2016.
\newblock Are emojis creating a new or old visual language for new generations?
  a socio-semiotic study.
\newblock \emph{Advances in Language and Literary Studies}, 7(6).

\bibitem[{Barbieri et~al.(2018{\natexlab{a}})Barbieri, Camacho-Collados,
  Ronzano, Espinosa-Anke, Ballesteros, Basile, Patti, and
  Saggion}]{barbieri-etal-2018-semeval}
Francesco Barbieri, Jose Camacho-Collados, Francesco Ronzano, Luis
  Espinosa-Anke, Miguel Ballesteros, Valerio Basile, Viviana Patti, and Horacio
  Saggion. 2018{\natexlab{a}}.
\newblock \href {https://doi.org/10.18653/v1/S18-1003} {{S}em{E}val 2018 task
  2: Multilingual emoji prediction}.
\newblock In \emph{Proceedings of The 12th International Workshop on Semantic
  Evaluation}, pages 24--33, New Orleans, Louisiana. Association for
  Computational Linguistics.

\bibitem[{Barbieri et~al.(2018{\natexlab{b}})Barbieri, Espinosa-Anke,
  Camacho-Collados, Schockaert, and Saggion}]{barbieri-etal-2018-interpretable}
Francesco Barbieri, Luis Espinosa-Anke, Jose Camacho-Collados, Steven
  Schockaert, and Horacio Saggion. 2018{\natexlab{b}}.
\newblock \href {https://doi.org/10.18653/v1/D18-1508} {Interpretable emoji
  prediction via label-wise attention {LSTM}s}.
\newblock In \emph{Proceedings of the 2018 Conference on Empirical Methods in
  Natural Language Processing}, pages 4766--4771, Brussels, Belgium.
  Association for Computational Linguistics.

\bibitem[{Biewald(2020)}]{wandb}
Lukas Biewald. 2020.
\newblock \href {https://www.wandb.com/} {Experiment tracking with weights and
  biases}.
\newblock Software available from wandb.com.

\bibitem[{Buda et~al.(2018)Buda, Maki, and Mazurowski}]{buda2018systematic}
Mateusz Buda, Atsuto Maki, and Maciej~A Mazurowski. 2018.
\newblock A systematic study of the class imbalance problem in convolutional
  neural networks.
\newblock \emph{Neural Networks}, 106:249--259.

\bibitem[{Cappallo et~al.(2018)Cappallo, Svetlichnaya, Garrigues, Mensink, and
  Snoek}]{cappallo2018new}
Spencer Cappallo, Stacey Svetlichnaya, Pierre Garrigues, Thomas Mensink, and
  Cees~GM Snoek. 2018.
\newblock New modality: Emoji challenges in prediction, anticipation, and
  retrieval.
\newblock \emph{IEEE Transactions on Multimedia}, 21(2):402--415.

\bibitem[{Chawla et~al.(2002)Chawla, Bowyer, Hall, and
  Kegelmeyer}]{chawla2002smote}
Nitesh~V Chawla, Kevin~W Bowyer, Lawrence~O Hall, and W~Philip Kegelmeyer.
  2002.
\newblock Smote: synthetic minority over-sampling technique.
\newblock \emph{Journal of artificial intelligence research}, 16:321--357.

\bibitem[{Choudhary et~al.(2018{\natexlab{a}})Choudhary, Singh, Anvesh~Rao, and
  Shrivastava}]{choudhary-etal-2018-twitter}
Nurendra Choudhary, Rajat Singh, Vijjini Anvesh~Rao, and Manish Shrivastava.
  2018{\natexlab{a}}.
\newblock \href {https://aclanthology.org/C18-1133} {{T}witter corpus of
  resource-scarce languages for sentiment analysis and multilingual emoji
  prediction}.
\newblock In \emph{Proceedings of the 27th International Conference on
  Computational Linguistics}, pages 1570--1577, Santa Fe, New Mexico, USA.
  Association for Computational Linguistics.

\bibitem[{Choudhary et~al.(2018{\natexlab{b}})Choudhary, Singh, Bindlish, and
  Shrivastava}]{choudhary2018contrastive}
Nurendra Choudhary, Rajat Singh, Ishita Bindlish, and Manish Shrivastava.
  2018{\natexlab{b}}.
\newblock \href {https://arxiv.org/pdf/1804.01855} {Contrastive learning of
  emoji-based representations for resource-poor languages}.
\newblock \emph{arXiv preprint arXiv:1804.01855}.

\bibitem[{Conneau et~al.(2020)Conneau, Khandelwal, Goyal, Chaudhary, Wenzek,
  Guzm{\'a}n, Grave, Ott, Zettlemoyer, and
  Stoyanov}]{conneau-etal-2020-unsupervised}
Alexis Conneau, Kartikay Khandelwal, Naman Goyal, Vishrav Chaudhary, Guillaume
  Wenzek, Francisco Guzm{\'a}n, Edouard Grave, Myle Ott, Luke Zettlemoyer, and
  Veselin Stoyanov. 2020.
\newblock \href {https://doi.org/10.18653/v1/2020.acl-main.747} {Unsupervised
  cross-lingual representation learning at scale}.
\newblock In \emph{Proceedings of the 58th Annual Meeting of the Association
  for Computational Linguistics}, pages 8440--8451, Online. Association for
  Computational Linguistics.

\bibitem[{Devlin et~al.(2019)Devlin, Chang, Lee, and
  Toutanova}]{devlin-etal-2019-bert}
Jacob Devlin, Ming-Wei Chang, Kenton Lee, and Kristina Toutanova. 2019.
\newblock \href {https://doi.org/10.18653/v1/N19-1423} {{BERT}: Pre-training of
  deep bidirectional transformers for language understanding}.
\newblock In \emph{Proceedings of the 2019 Conference of the North {A}merican
  Chapter of the Association for Computational Linguistics: Human Language
  Technologies, Volume 1 (Long and Short Papers)}, pages 4171--4186,
  Minneapolis, Minnesota. Association for Computational Linguistics.

\bibitem[{Drumnond(2003)}]{drumnond2003class}
Chris Drumnond. 2003.
\newblock Class imbalance and cost sensitivity: Why undersampling beats
  oversampling.
\newblock In \emph{ICML-KDD 2003 Workshop: Learning from Imbalanced Datasets},
  volume~3.

\bibitem[{Duarte et~al.(2020)Duarte, Macedo, and
  Gon{\c{c}}alo~Oliveira}]{duarte2020emoji}
Luis Duarte, Lu{\'\i}s Macedo, and Hugo Gon{\c{c}}alo~Oliveira. 2020.
\newblock Emoji prediction for portuguese.
\newblock In \emph{International Conference on Computational Processing of the
  Portuguese Language}, pages 174--183. Springer.

\bibitem[{Felbo et~al.(2017)Felbo, Mislove, S{\o}gaard, Rahwan, and
  Lehmann}]{felbo-etal-2017-using}
Bjarke Felbo, Alan Mislove, Anders S{\o}gaard, Iyad Rahwan, and Sune Lehmann.
  2017.
\newblock \href {https://doi.org/10.18653/v1/D17-1169} {Using millions of emoji
  occurrences to learn any-domain representations for detecting sentiment,
  emotion and sarcasm}.
\newblock In \emph{Proceedings of the 2017 Conference on Empirical Methods in
  Natural Language Processing}, pages 1615--1625, Copenhagen, Denmark.
  Association for Computational Linguistics.

\bibitem[{Francis et~al.(2021)Francis, Tenison, and Rish}]{francis2021towards}
Sreya Francis, Irene Tenison, and Irina Rish. 2021.
\newblock Towards causal federated learning for enhanced robustness and
  privacy.
\newblock \emph{arXiv preprint arXiv:2104.06557}.

\bibitem[{Geifman and El-Yaniv(2017)}]{geifman2017deep}
Yonatan Geifman and Ran El-Yaniv. 2017.
\newblock Deep active learning over the long tail.
\newblock \emph{arXiv preprint arXiv:1711.00941}.

\bibitem[{He and Garcia(2009)}]{he2009learning}
Haibo He and Edwardo~A Garcia. 2009.
\newblock Learning from imbalanced data.
\newblock \emph{IEEE Transactions on knowledge and data engineering},
  21(9):1263--1284.

\bibitem[{Hochreiter and Schmidhuber(1997)}]{Hochreiter_Schmidhuber_1997}
Sepp Hochreiter and Jürgen Schmidhuber. 1997.
\newblock \href {https://doi.org/10.1162/neco.1997.9.8.1735} {Long short-term
  memory}.
\newblock \emph{Neural Computation}, 9(8):1735–1780.

\bibitem[{Huang et~al.(2016)Huang, Li, Loy, and Tang}]{huang2016learning}
Chen Huang, Yining Li, Chen~Change Loy, and Xiaoou Tang. 2016.
\newblock Learning deep representation for imbalanced classification.
\newblock In \emph{Proceedings of the IEEE conference on computer vision and
  pattern recognition}, pages 5375--5384.

\bibitem[{Kakwani et~al.(2020)Kakwani, Kunchukuttan, Golla, N.C.,
  Bhattacharyya, Khapra, and Kumar}]{kakwani-etal-2020-indicnlpsuite}
Divyanshu Kakwani, Anoop Kunchukuttan, Satish Golla, Gokul N.C., Avik
  Bhattacharyya, Mitesh~M. Khapra, and Pratyush Kumar. 2020.
\newblock \href {https://doi.org/10.18653/v1/2020.findings-emnlp.445}
  {{I}ndic{NLPS}uite: Monolingual corpora, evaluation benchmarks and
  pre-trained multilingual language models for {I}ndian languages}.
\newblock In \emph{Findings of the Association for Computational Linguistics:
  EMNLP 2020}, pages 4948--4961, Online. Association for Computational
  Linguistics.

\bibitem[{Khan et~al.(2017)Khan, Hayat, Bennamoun, Sohel, and
  Togneri}]{khan2017cost}
Salman~H Khan, Munawar Hayat, Mohammed Bennamoun, Ferdous~A Sohel, and Roberto
  Togneri. 2017.
\newblock Cost-sensitive learning of deep feature representations from
  imbalanced data.
\newblock \emph{IEEE transactions on neural networks and learning systems},
  29(8):3573--3587.

\bibitem[{Li et~al.(2018)Li, Sahu, Zaheer, Sanjabi, Talwalkar, and
  Smith}]{li2018federated}
Tian Li, Anit~Kumar Sahu, Manzil Zaheer, Maziar Sanjabi, Ameet Talwalkar, and
  Virginia Smith. 2018.
\newblock Federated optimization in heterogeneous networks.
\newblock \emph{arXiv preprint arXiv:1812.06127}.

\bibitem[{Liebeskind and Liebeskind(2019)}]{chayahebrew}
Chaya Liebeskind and Shmuel Liebeskind. 2019.
\newblock \href {https://doi.org/10.1145/3308560.3316548} {Emoji prediction for
  hebrew political domain}.
\newblock In \emph{Companion Proceedings of The 2019 World Wide Web
  Conference}, WWW '19, page 468–477, New York, NY, USA. Association for
  Computing Machinery.

\bibitem[{Lin et~al.(2017)Lin, Goyal, Girshick, He, and
  Doll{\'a}r}]{lin2017focal}
Tsung-Yi Lin, Priya Goyal, Ross Girshick, Kaiming He, and Piotr Doll{\'a}r.
  2017.
\newblock Focal loss for dense object detection.
\newblock In \emph{Proceedings of the IEEE international conference on computer
  vision}, pages 2980--2988.

\bibitem[{Ma et~al.(2020)Ma, Liu, Wang, and Vosoughi}]{ma2020emoji}
Weicheng Ma, Ruibo Liu, Lili Wang, and Soroush Vosoughi. 2020.
\newblock Emoji prediction: Extensions and benchmarking.
\newblock \emph{arXiv preprint arXiv:2007.07389}.

\bibitem[{McMahan et~al.(2017)McMahan, Moore, Ramage, Hampson, and
  y~Arcas}]{mcmahan2017communication}
Brendan McMahan, Eider Moore, Daniel Ramage, Seth Hampson, and Blaise~Aguera
  y~Arcas. 2017.
\newblock \href {http://proceedings.mlr.press/v54/mcmahan17a/mcmahan17a.pdf}
  {Communication-efficient learning of deep networks from decentralized data}.
\newblock In \emph{Artificial intelligence and statistics}, pages 1273--1282.
  PMLR.

\bibitem[{Padilla~L{\'o}pez and Cap(2017)}]{padilla-lopez-cap-2017-ever}
Rebeca Padilla~L{\'o}pez and Fabienne Cap. 2017.
\newblock \href {https://doi.org/10.18653/v1/W17-5215} {Did you ever read about
  frogs drinking coffee? investigating the compositionality of multi-emoji
  expressions}.
\newblock In \emph{Proceedings of the 8th Workshop on Computational Approaches
  to Subjectivity, Sentiment and Social Media Analysis}, pages 113--117,
  Copenhagen, Denmark. Association for Computational Linguistics.

\bibitem[{Paszke et~al.(2019)Paszke, Gross, Massa, Lerer, Bradbury, Chanan,
  Killeen, Lin, Gimelshein, Antiga, and et~al.}]{Paszke_et_al}
Adam Paszke, Sam Gross, Francisco Massa, Adam Lerer, James Bradbury, Gregory
  Chanan, Trevor Killeen, Zeming Lin, Natalia Gimelshein, Luca Antiga, and
  et~al. 2019.
\newblock \href
  {http://papers.neurips.cc/paper/9015-pytorch-an-imperative-style-high-performance-deep-learning-library.pdf}
  {\emph{PyTorch: An Imperative Style, High-Performance Deep Learning
  Library}}, page 8024–8035. Curran Associates, Inc.

\bibitem[{Ramaswamy et~al.(2019)Ramaswamy, Mathews, Rao, and
  Beaufays}]{ramaswamy2019federated}
Swaroop Ramaswamy, Rajiv Mathews, Kanishka Rao, and Fran{\c{c}}oise Beaufays.
  2019.
\newblock Federated learning for emoji prediction in a mobile keyboard.
\newblock \emph{arXiv preprint arXiv:1906.04329}.

\bibitem[{Ronzano et~al.(2018)Ronzano, Barbieri, Wahyu~Pamungkas, Patti,
  Chiusaroli et~al.}]{ronzano2018overview}
Francesco Ronzano, Francesco Barbieri, Endang Wahyu~Pamungkas, Viviana Patti,
  Francesca Chiusaroli, et~al. 2018.
\newblock Overview of the evalita 2018 italian emoji prediction (itamoji) task.
\newblock In \emph{6th Evaluation Campaign of Natural Language Processing and
  Speech Tools for Italian. Final Workshop, EVALITA 2018}, volume 2263, pages
  1--9. CEUR-WS.

\bibitem[{Sarafianos et~al.(2018)Sarafianos, Xu, and
  Kakadiaris}]{sarafianos2018deep}
Nikolaos Sarafianos, Xiang Xu, and Ioannis~A Kakadiaris. 2018.
\newblock Deep imbalanced attribute classification using visual attention
  aggregation.
\newblock In \emph{Proceedings of the European Conference on Computer Vision
  (ECCV)}, pages 680--697.

\bibitem[{Shen et~al.(2016)Shen, Lin, and Huang}]{shen2016relay}
Li~Shen, Zhouchen Lin, and Qingming Huang. 2016.
\newblock Relay backpropagation for effective learning of deep convolutional
  neural networks.
\newblock In \emph{European conference on computer vision}, pages 467--482.
  Springer.

\bibitem[{Stark and Crawford(2015)}]{stark2015conservatism}
Luke Stark and Kate Crawford. 2015.
\newblock The conservatism of emoji: Work, affect, and communication.
\newblock \emph{Social Media+ Society}, 1(2):2056305115604853.

\bibitem[{Ting(2000)}]{ting2000comparative}
Kai~Ming Ting. 2000.
\newblock A comparative study of cost-sensitive boosting algorithms.
\newblock In \emph{In Proceedings of the 17th International Conference on
  Machine Learning}. Citeseer.

\bibitem[{Tomihira et~al.(2020)Tomihira, Otsuka, Yamashita, and
  Satoh}]{tomihira2020multilingual}
Toshiki Tomihira, Atsushi Otsuka, Akihiro Yamashita, and Tetsuji Satoh. 2020.
\newblock \href
  {https://www.emerald.com/insight/content/doi/10.1108/IJWIS-09-2019-0042/full/html}
  {Multilingual emoji prediction using bert for sentiment analysis}.
\newblock \emph{International Journal of Web Information Systems}.

\bibitem[{Wolf et~al.(2020)Wolf, Debut, Sanh, Chaumond, Delangue, Moi, Cistac,
  Rault, Louf, Funtowicz, Davison, Shleifer, von Platen, Ma, Jernite, Plu, Xu,
  Scao, Gugger, Drame, Lhoest, and Rush}]{wolf-etal-2020-transformers}
Thomas Wolf, Lysandre Debut, Victor Sanh, Julien Chaumond, Clement Delangue,
  Anthony Moi, Pierric Cistac, Tim Rault, Rémi Louf, Morgan Funtowicz, Joe
  Davison, Sam Shleifer, Patrick von Platen, Clara Ma, Yacine Jernite, Julien
  Plu, Canwen Xu, Teven~Le Scao, Sylvain Gugger, Mariama Drame, Quentin Lhoest,
  and Alexander~M. Rush. 2020.
\newblock \href {https://www.aclweb.org/anthology/2020.emnlp-demos.6}
  {Transformers: State-of-the-art natural language processing}.
\newblock In \emph{Proceedings of the 2020 Conference on Empirical Methods in
  Natural Language Processing: System Demonstrations}, pages 38--45, Online.
  Association for Computational Linguistics.

\bibitem[{Yang et~al.(2018)Yang, Andrew, Eichner, Sun, Li, Kong, Ramage, and
  Beaufays}]{yang2018applied}
Timothy Yang, Galen Andrew, Hubert Eichner, Haicheng Sun, Wei Li, Nicholas
  Kong, Daniel Ramage, and Fran{\c{c}}oise Beaufays. 2018.
\newblock Applied federated learning: Improving google keyboard query
  suggestions.
\newblock \emph{arXiv preprint arXiv:1812.02903}.

\bibitem[{Zhang et~al.(2020)Zhang, Zhou, Erekhinskaya, and
  Moldovan}]{zhang2020emoji}
Linrui Zhang, Yisheng Zhou, Tatiana Erekhinskaya, and Dan Moldovan. 2020.
\newblock \href
  {https://link.springer.com/chapter/10.1007/978-3-030-39442-4_65} {Emoji
  prediction: A transfer learning approach}.
\newblock In \emph{Future of Information and Communication Conference}, pages
  864--872. Springer.

\bibitem[{Zhou and Liu(2005)}]{zhou2005training}
Zhi-Hua Zhou and Xu-Ying Liu. 2005.
\newblock Training cost-sensitive neural networks with methods addressing the
  class imbalance problem.
\newblock \emph{IEEE Transactions on knowledge and data engineering},
  18(1):63--77.

\bibitem[{Zou et~al.(2018)Zou, Yu, Kumar, and Wang}]{zou2018unsupervised}
Yang Zou, Zhiding Yu, BVK Kumar, and Jinsong Wang. 2018.
\newblock Unsupervised domain adaptation for semantic segmentation via
  class-balanced self-training.
\newblock In \emph{Proceedings of the European conference on computer vision
  (ECCV)}, pages 289--305.

\end{thebibliography}
\bibliographystyle{acl_natbib}

\appendix

\section{Appendix}
\label{sec:appendix}

\subsection{Data}
\label{sec:appendix_data}

The tweets were collected using an "Elevated access" to the Twitter API v2. 
To collect tweets written in the Hindi language, we use "lang:hi" query. 
No other search criteria is used. 
The time-span of the tweets is from 19th April, 2021 to 8th May, 2021.
Figure~\ref{fig:example_dataset} shows a sample of tweets present in our Hindi dataset for the task of emoji prediction.

\begin{figure}[ht]
\centering
\includegraphics[width=\columnwidth]{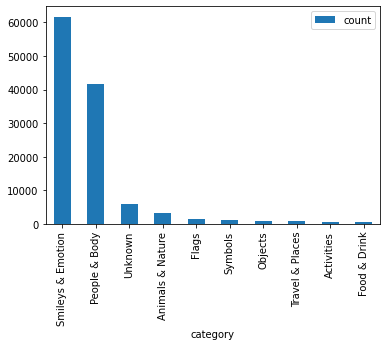}
\caption{Category distribution of complete dataset} 
\label{fig3}
\end{figure}

\begin{figure*}
\centering
\includegraphics[width=\textwidth]{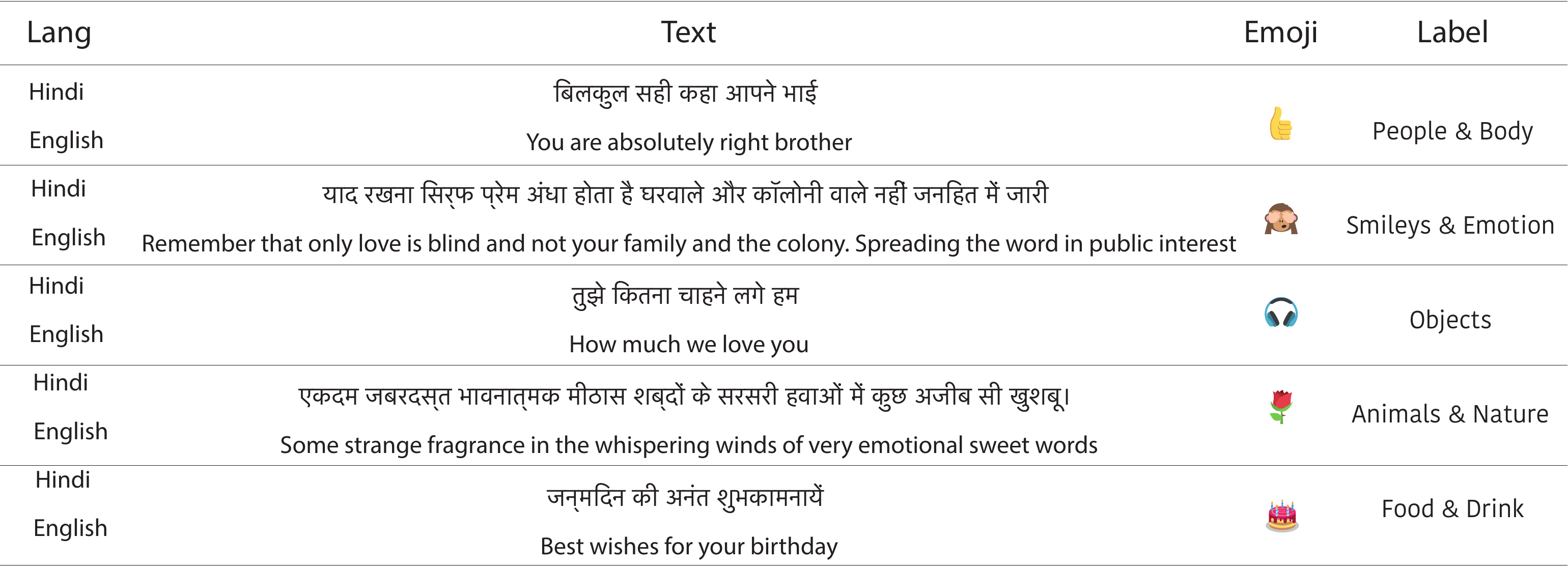}
\caption{Example of our Hindi dataset}
\label{fig:example_dataset}
\end{figure*}

\subsection{Server-Based Models}
\label{sec:appendix_server}

For traditional server-side transformer models, we use the simple transformers library.\footnote{https://simpletransformers.ai/} 
We use the default configuration options and train all the transformer models for 25 epochs with a learning rate of 4e-5 and no weight decay or momentum. 

All baseline models (transformer-based and others) are trained with batch size $8$, learning rate $4e-5$, and seq. length $128$.

All the models were trained using deterministic algorithms for randomness\footnote{https://pytorch.org/docs/stable/notes/randomness.html} in PyTorch and are easily reproducible using the same seeds.

\subsection{Experiments considering only text}
\label{sec:additional_exps}
We run some additional experiments considering only the actually text without applying the token setup as described in \S\ref{sec:Dataset and Pre-processing} as reflected in Tables~\ref{og_fed_results}, \ref{og_shared_fed_results}, \ref{og_causalFedGSD}, and \ref{og_result_comparison}. 
Observing these results, we note that except for non-I.I.D. settings, we see negligible improvements from applying the extra tokens in the original dataset (see \S\ref{sec:Dataset and Pre-processing}).
\begin{table*}[]
\resizebox{\textwidth}{!}{
\begin{tabular}{c|ccc|ccc|ccc|ccc|ccc|ccc}

 & \multicolumn{6}{c|}{$\textbf{c = 10\%}$} 
 & \multicolumn{6}{c|}{$\textbf{c = 30\%}$} 
 & \multicolumn{6}{c}{$\textbf{c = 50\%}$} \\\hline
 & \multicolumn{3}{c|}{IID} & \multicolumn{3}{c|}{non-IID} 
 & \multicolumn{3}{c|}{IID} & \multicolumn{3}{c|}{non-IID} 
 & \multicolumn{3}{c|}{IID} & \multicolumn{3}{c}{non-IID} \\  
 & Precision    & Recall    & F1    & Precision      & Recall     & F1     & Precision    & Recall    & F1    & Precision      & Recall     & F1     & Precision    & Recall    & F1    & Precision      & Recall     & F1     \\ \hline
\textbf{Imbalanced} &
  $61.33$ &
  $64.66$ &
  $62.32$ &
  $57.70$ &
  $64.10$ &
  $57.96$ &
  $61.55$ &
  $\textbf{67.64}$ &
  $63.60$ &
  $58.01$ &
  $58.42$ &
  $54.86$ &
  $61.65$ &
  $66.83$ &
  $63.57$ &
  ${58.30}$ &
  $61.59$ &
  $58.09$ \\ 
\textbf{Re-sampled} &
  $61.49$ &
  $46.22$ &
  $51.12$ &
  $56.84$ &
  $30.06$ &
  $34.28$ &
  $60.60$ &
  $43.75$ &
  $49.19$ &
  $57.48$ &
  $35.32$ &
  $41.36$ &
  $60.85$ &
  $47.71$ &
  $52.14$ &
  $56.13$ &
  $41.28$ &
  $45.76$ \\ 
\textbf{Cost-Sensitive} &
  $62.14$ &
  $63.35$ &
  $61.99$ &
  $58.08$ &
  ${65.86}$ &
  ${61.25}$ &
  $\textbf{63.72}$ &
  $65.25$ &
  $\textbf{63.78}$ &
  $56.39$ &
  $57.76$ &
  $54.36$ &
  $60.36$ &
  $59.99$ &
  $59.57$ &
  $56.68$ &
  $63.22$ &
  $59.36$ \\ 
\end{tabular}
}
\caption{Results using the FedProx algorithm on the dataset as explained in~\ref{sec:additional_exps}.}
\label{og_fed_results}
\end{table*}

\begin{table*}[!h]
\resizebox{\textwidth}{!}{
\begin{tabular}{c|ccc|ccc|ccc|ccc|ccc|ccc}

 & \multicolumn{6}{c|}{$\textbf{c = 10\%}$} 
 & \multicolumn{6}{c|}{$\textbf{c = 30\%}$} 
 & \multicolumn{6}{c}{$\textbf{c = 50\%}$} \\\hline
 & \multicolumn{3}{c|}{IID} & \multicolumn{3}{c|}{non-IID} 
 & \multicolumn{3}{c|}{IID} & \multicolumn{3}{c|}{non-IID} 
 & \multicolumn{3}{c|}{IID} & \multicolumn{3}{c}{non-IID} \\  
 & Precision    & Recall    & F1    & Precision      & Recall     & F1     & Precision    & Recall    & F1    & Precision      & Recall     & F1     & Precision    & Recall    & F1    & Precision      & Recall     & F1     \\ \hline
\textbf{Imbalanced} &
  $\textbf{61.83}$ &
  $67.24$ &
  $\textbf{63.87}$ &
  $58.96$ &
  $45.88$ &
  $38.34$ &
  $61.62$ &
  $67.11$ &
  $63.41$ &
  $58.95$ &
  $63.80$ &
  $60.58$ &
  $61.66$ &
  $\textbf{67.38}$ &
  $63.70$ &
  ${59.46}$ &
  $49.39$ &
  $43.88$ \\ 
\textbf{Re-sampled} &
  $59.44$ &
  $37.53$ &
  $43.68$ &
  $53.10$ &
  $49.91$ &
  $41.50$ &
  $59.53$ &
  $41.06$ &
  $46.54$ &
  $58.61$ &
  $26.68$ &
  $32.45$ &
  $60.97$ &
  $39.02$ &
  $45.48$ &
  $57.70$ &
  $32.98$ &
  $39.71$ \\ 
\textbf{Cost-Sensitive} &
  $60.88$ &
  $59.38$ &
  $59.49$ &
  $54.82$ &
  $57.42$ &
  $46.17$ &
  $60.45$ &
  $60.71$ &
  $59.96$ &
  $59.05$ &
  ${66.52}$ &
  ${62.09}$ &
  $60.44$ &
  $61.41$ &
  $60.38$ &
  $58.69$ &
  $63.60$ &
  $60.11$ \\ 
\end{tabular}
}
\caption{Results using the modified CausalFedGSD on the dataset as explained in~\ref{sec:additional_exps}.}
\label{og_shared_fed_results}
\end{table*}

\begin{table}

\resizebox{\columnwidth}{!}{

\begin{tabular}{c|ccc|ccc|ccc}

& \multicolumn{3}{c|}{$\textbf{c = 10\%}$} 
& \multicolumn{3}{c|}{$\textbf{c = 30\%}$} 
& \multicolumn{3}{c}{$\textbf{c = 50\%}$} \\ 

& Precision & Recall & F1        
& Precision & Recall & F1             
& Precision & Recall & F1   \\ \hline

\textbf{Imbalanced}     
& $62.93$ & $63.37$ & $62.68$    
& $63.01$ & $63.48$ & $62.73$          
& $63.21$ & $63.60$ & $62.90$    \\ 

\textbf{Re-sampled}     
& $60.84$ & $49.59$ & $53.30$    
& $60.72$ & $49.45$ & $53.11$          
& $60.47$ & $49.08$ & $52.76$    \\ 

\textbf{Cost-Sensitive} 
& $64.56$ & $63.70$ & $63.65$    
& $\textbf{64.61}$ & $\textbf{63.88}$ & $\textbf{63.70}$ 
& $64.27$ & $63.41$ & $63.33$    \\ 

\end{tabular}
}

\caption{Results for the baseline CausalFedGSD on the dataset as explained in~\ref{sec:additional_exps}.}
\label{og_causalFedGSD}

\end{table}

\begin{table}
\resizebox{\columnwidth}{!}{
\begin{tabular}{lc|cc}
\hline \textbf{Approach} & Centralised & \multicolumn{2}{c} { Federated} \\
\hline & XLM-R & FedProx & Modified CausalFedGSD\\
\cline { 2 - 4 } 
Imbalanced & $69.44$ & $63.60$ & $63.87$\\
Re-sampled & $63.39$ & $52.14$ & $46.54$\\
Cost-Sensitive & $68.87$ & $63.78$ & $62.09$\\
\hline
\end{tabular}
}
\caption{An approach-wise comparison of F1 scores for best performing models in centralized and federated settings trained on the dataset as explained in~\ref{sec:additional_exps}.}
\label{og_result_comparison}
\end{table}

\subsection{Federated Learning Plots}
\label{sec:appendix_fed}
This section provides detailed graphs comparing the training loss, validation AUC, validation F1 score and validation accuracy for every dataset variation. All of these graphs were made using Weights and Biases~\citep{wandb}.

We set the value of the proximal term to $0.01$ following~\citet{li2018federated}. We set the learning rate as $1e-02$ for Federated models based on hyperparameter sweeps.

\subsubsection{Imbalanced Dataset (IID)}

\begin{figure}[H]
\centering
\includegraphics[width=0.8\columnwidth]{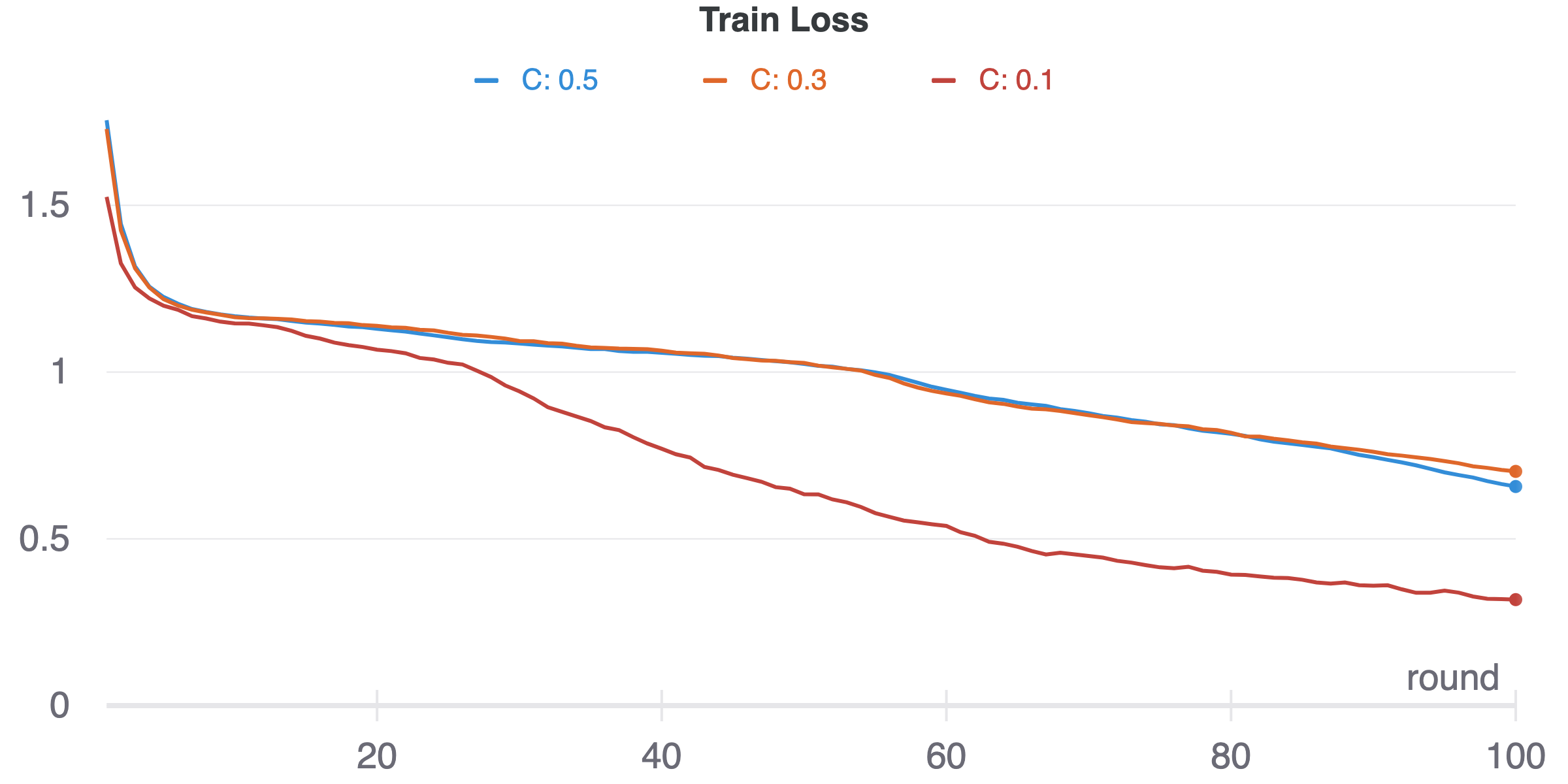}
\end{figure}
\vspace{2pt}

\begin{figure}[H]
\centering
\includegraphics[width=0.8\columnwidth]{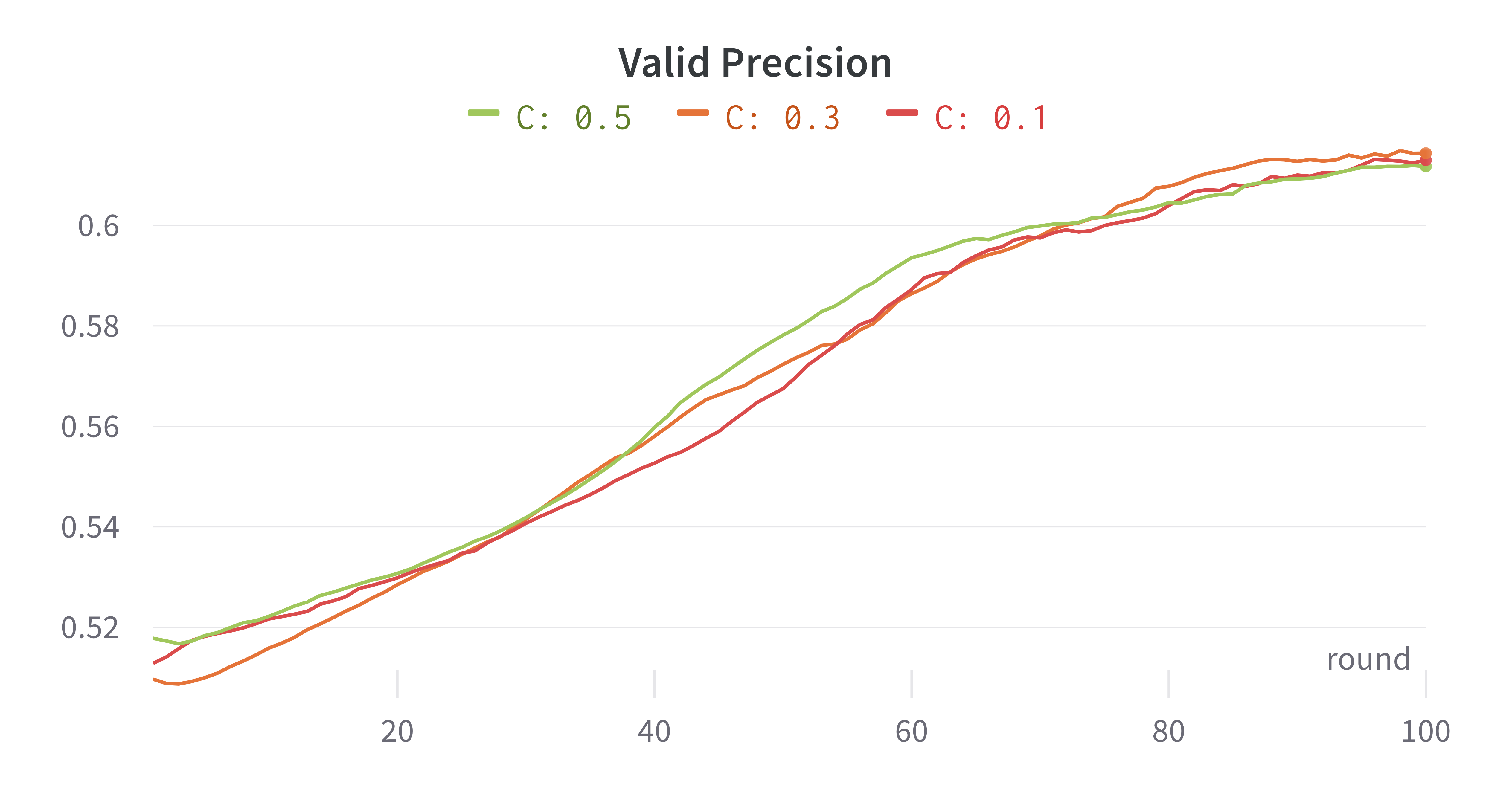}
\end{figure}
\vspace{2pt}

\begin{figure}[H]
\centering
\includegraphics[width=0.8\columnwidth]{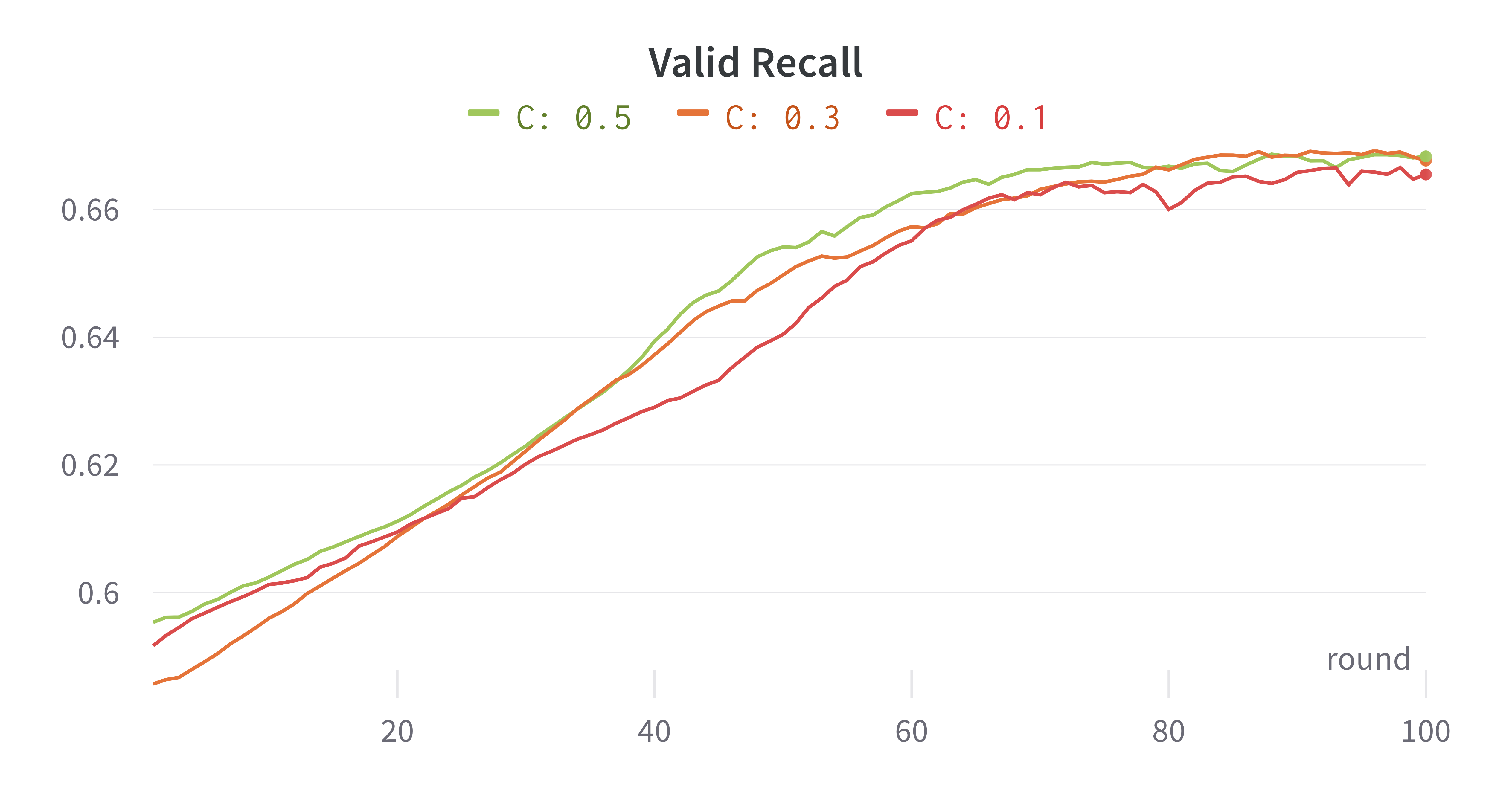}
\end{figure}
\vspace{2pt}

\begin{figure}[H]
\centering
\includegraphics[width=0.8\columnwidth]{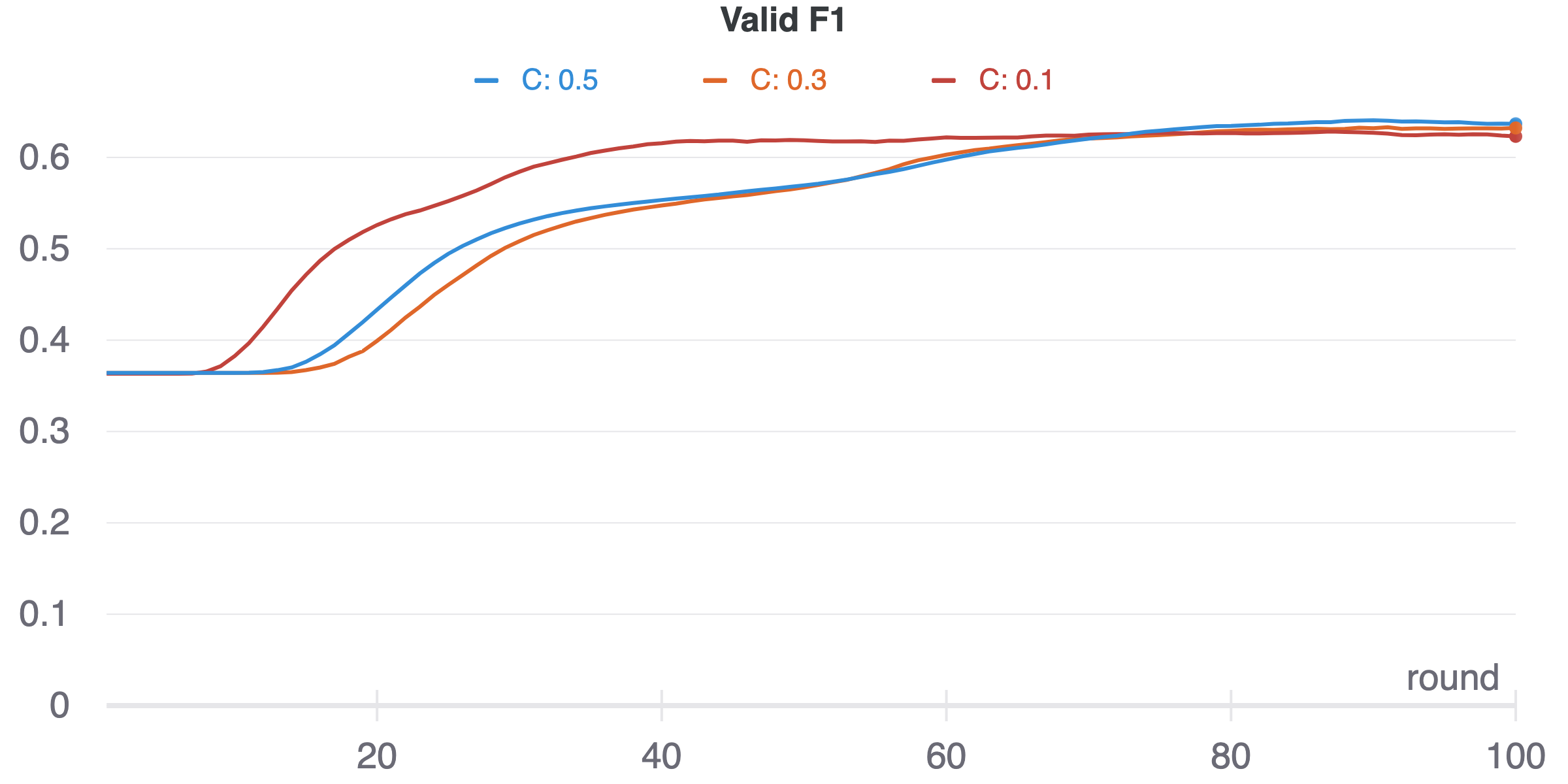}
\end{figure}
\vspace{2pt}

\subsubsection{Imbalanced Dataset (non-IID)}

\begin{figure}[H]
\centering
\includegraphics[width=0.8\columnwidth]{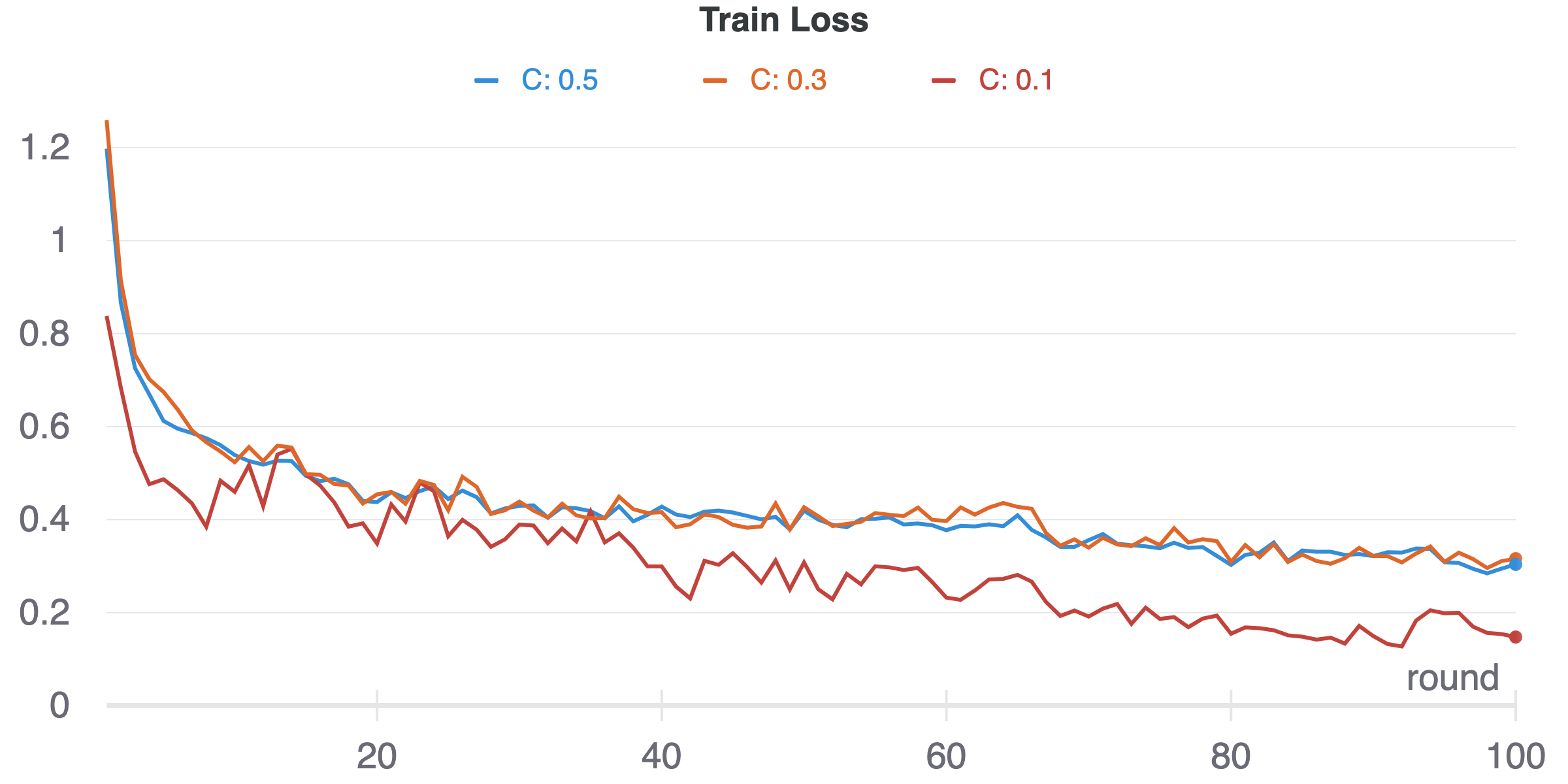}
\end{figure}
\vspace{2pt}

\begin{figure}[H]
\centering
\includegraphics[width=0.8\columnwidth]{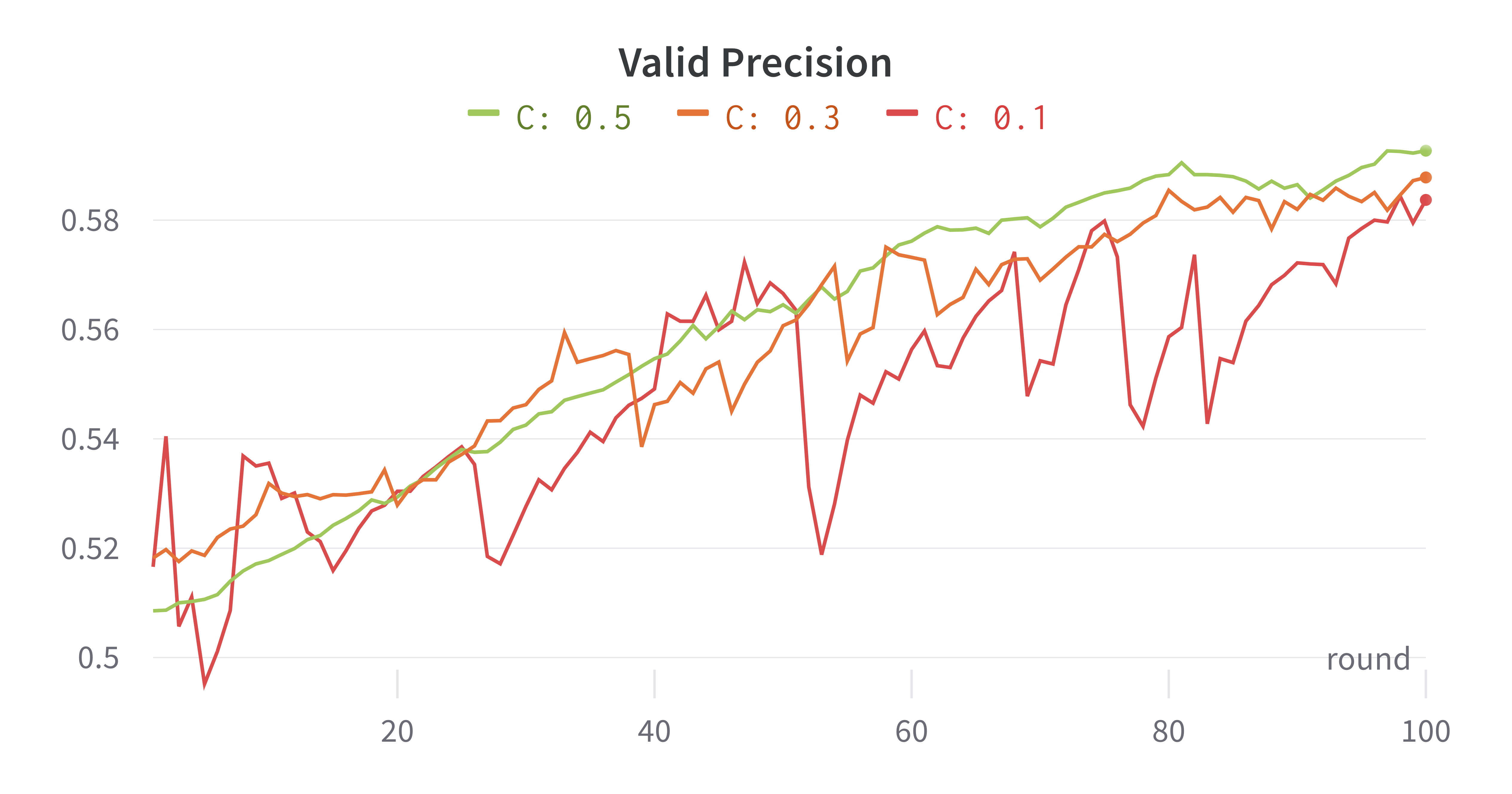}
\end{figure}
\vspace{2pt}

\begin{figure}[H]
\centering
\includegraphics[width=0.8\columnwidth]{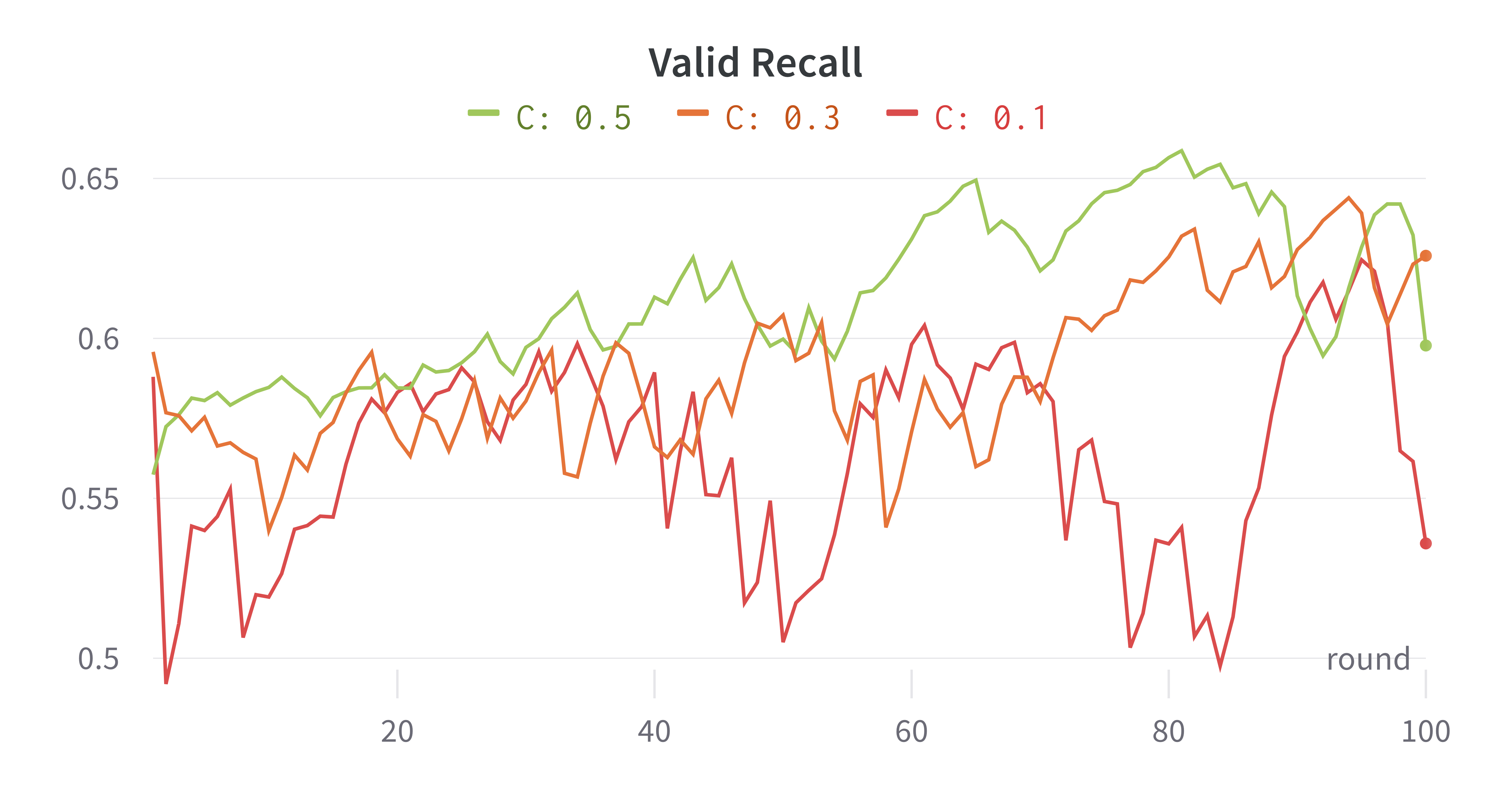}
\end{figure}
\vspace{2pt}

\begin{figure}[H]
\centering
\includegraphics[width=0.8\columnwidth]{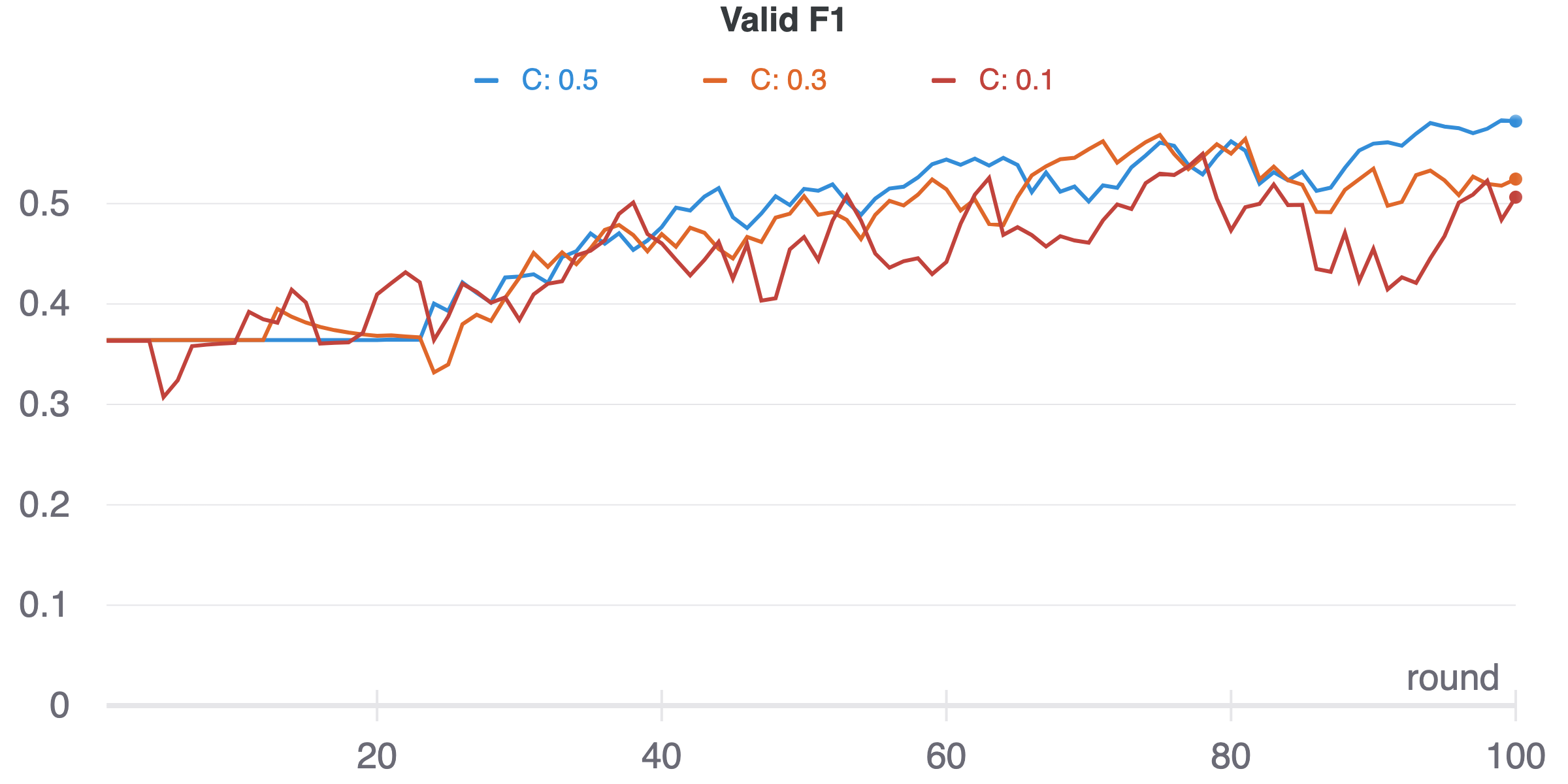}
\end{figure}
\vspace{2pt}

\subsubsection{Balanced Dataset (IID)}

\begin{figure}[H]
\centering
\includegraphics[width=0.8\columnwidth]{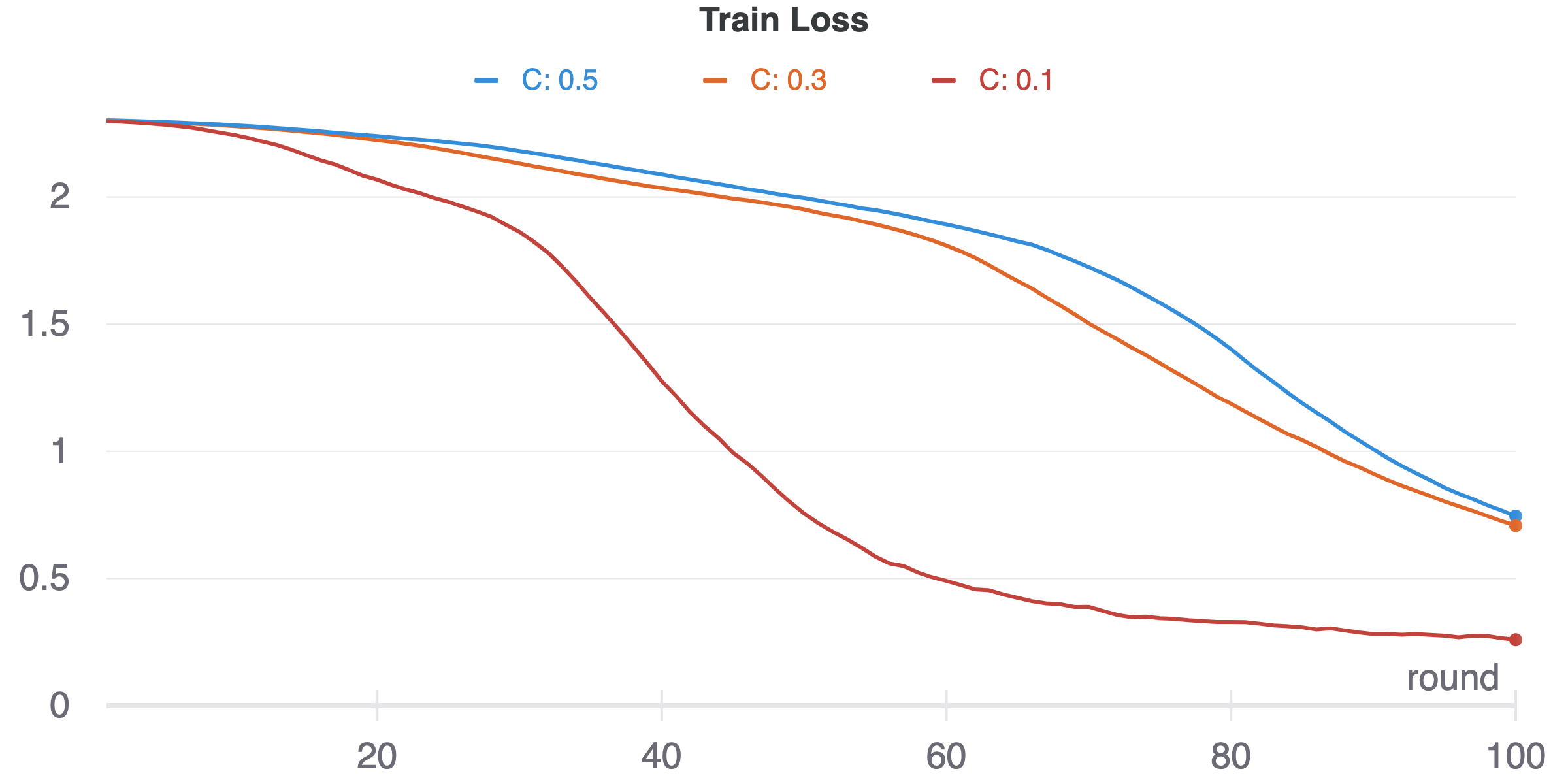}
\end{figure}
\vspace{2pt}

\begin{figure}[H]
\centering
\includegraphics[width=0.8\columnwidth]{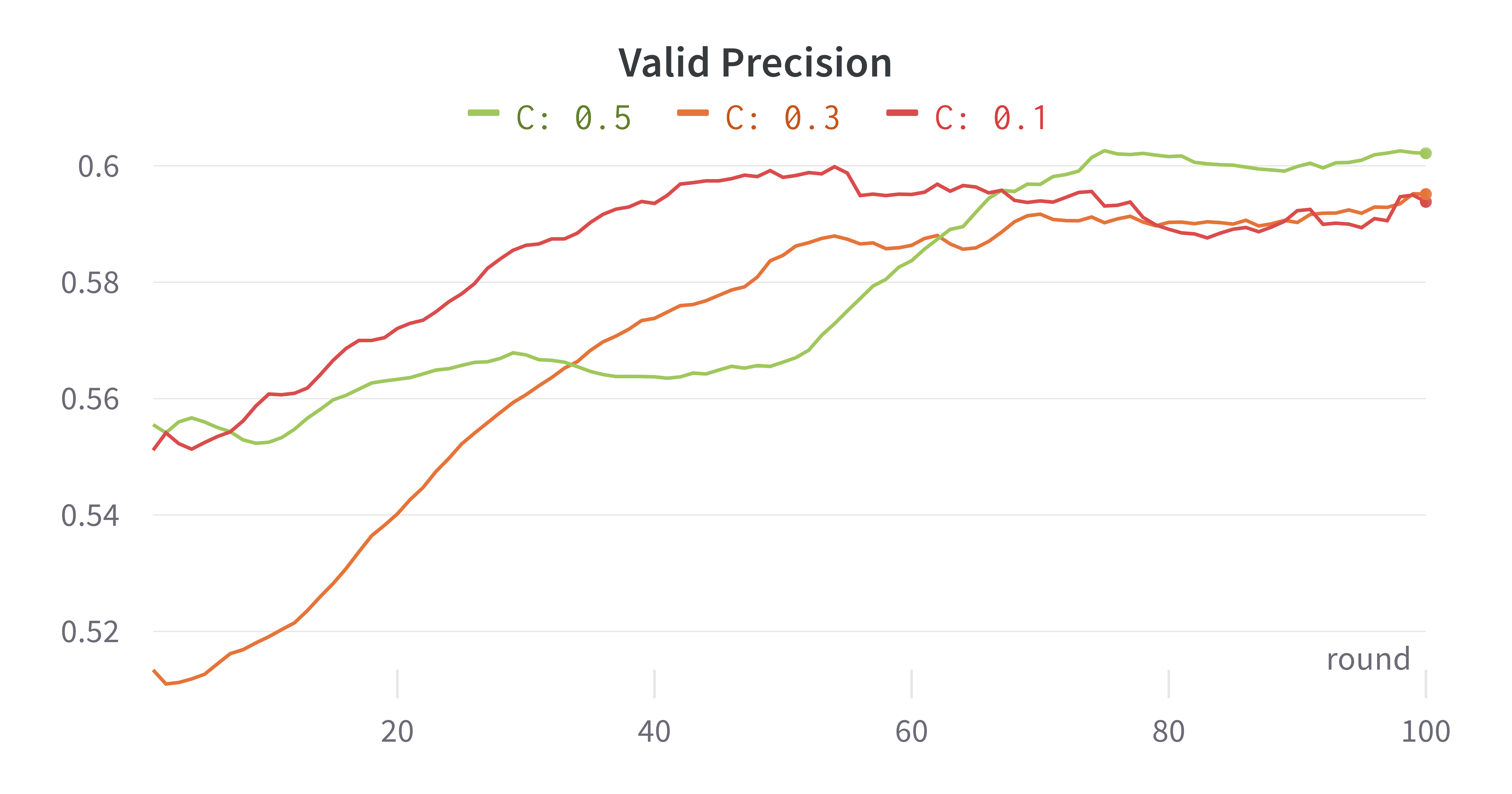}
\end{figure}
\vspace{2pt}

\begin{figure}[H]
\centering
\includegraphics[width=0.8\columnwidth]{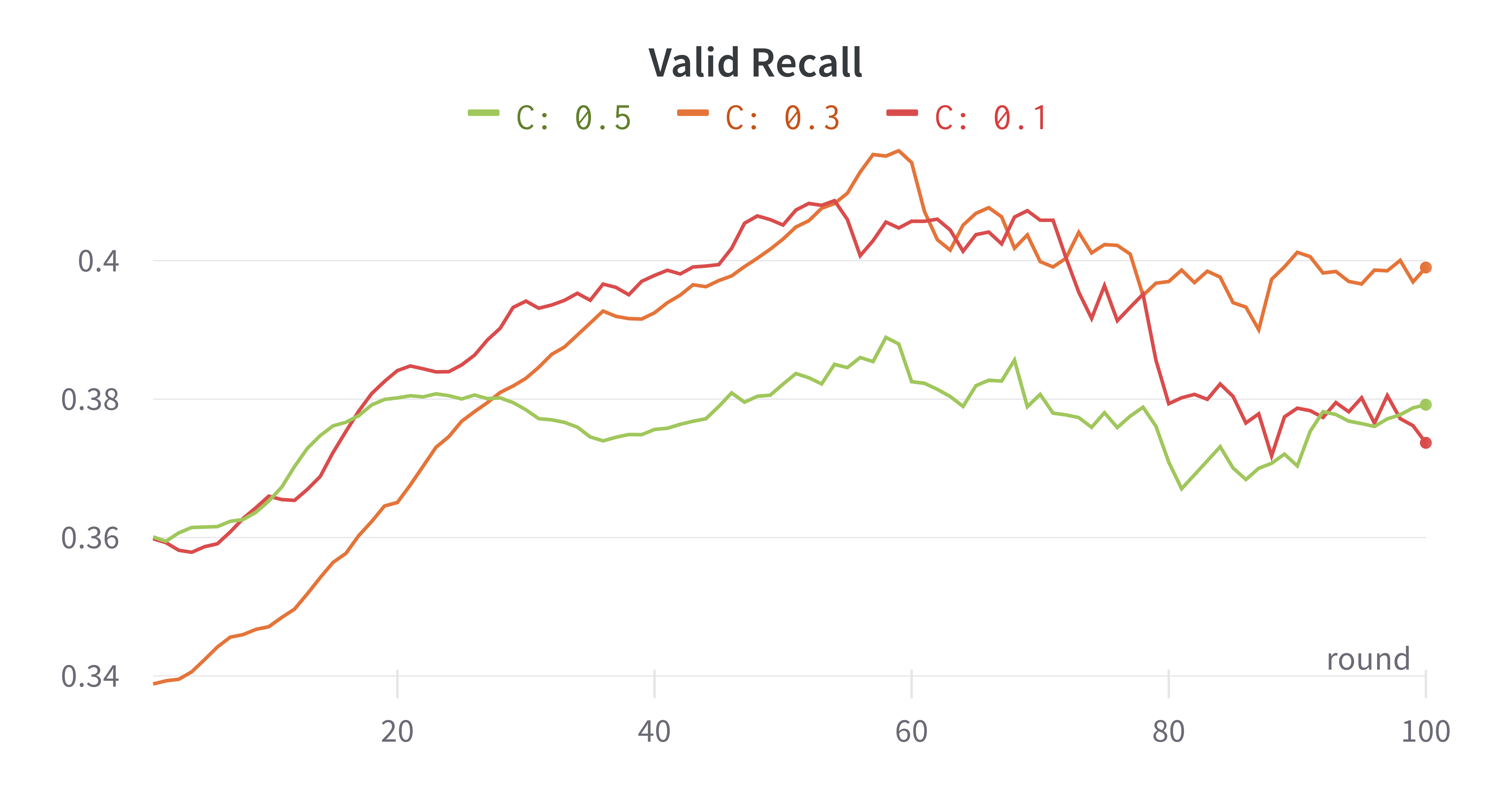}
\end{figure}
\vspace{2pt}

\begin{figure}[H]
\centering
\includegraphics[width=0.8\columnwidth]{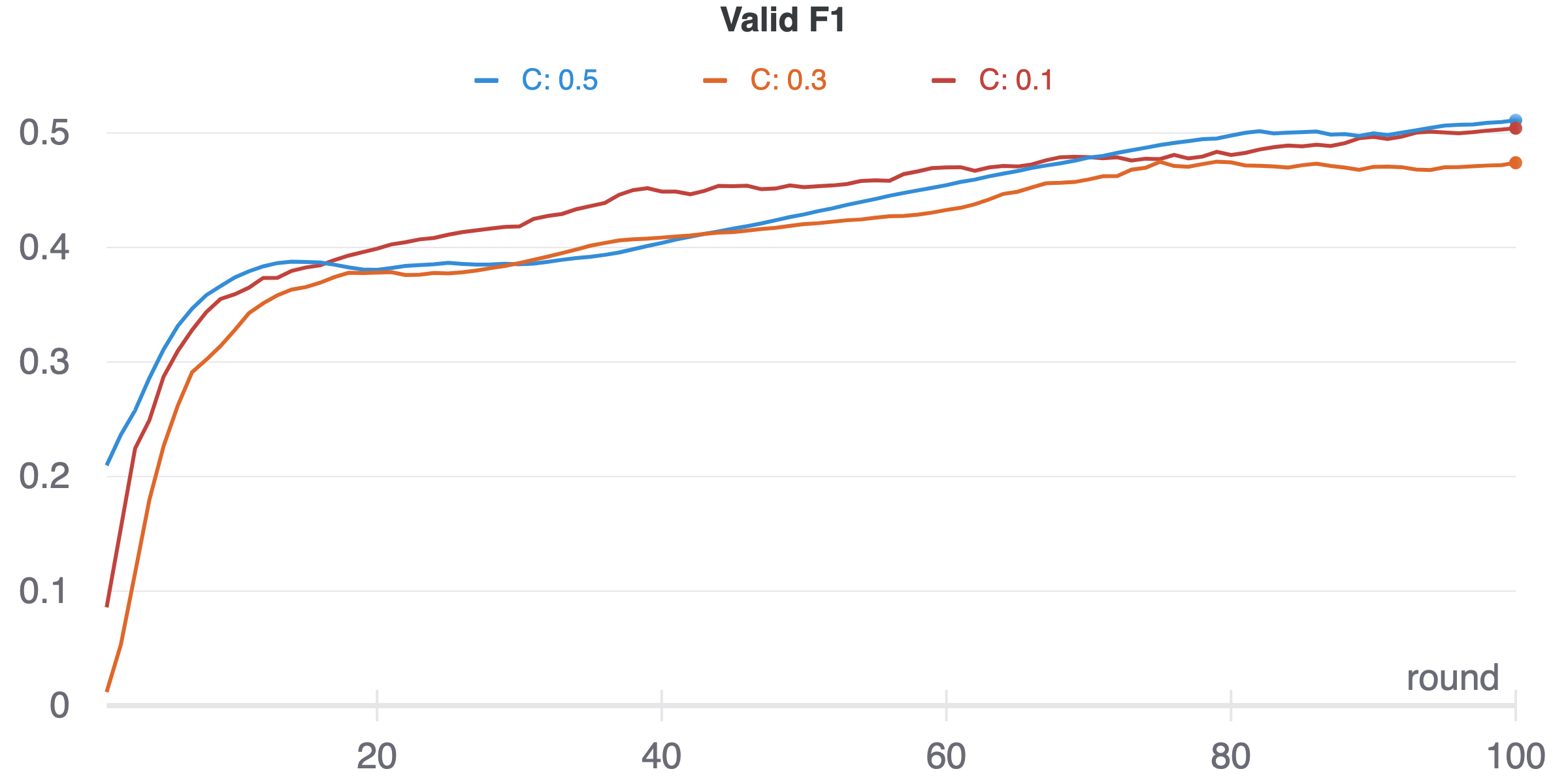}
\end{figure}
\vspace{2pt}

\subsubsection{Balanced Dataset (non-IID)}

\begin{figure}[H]
\centering
\includegraphics[width=0.8\columnwidth]{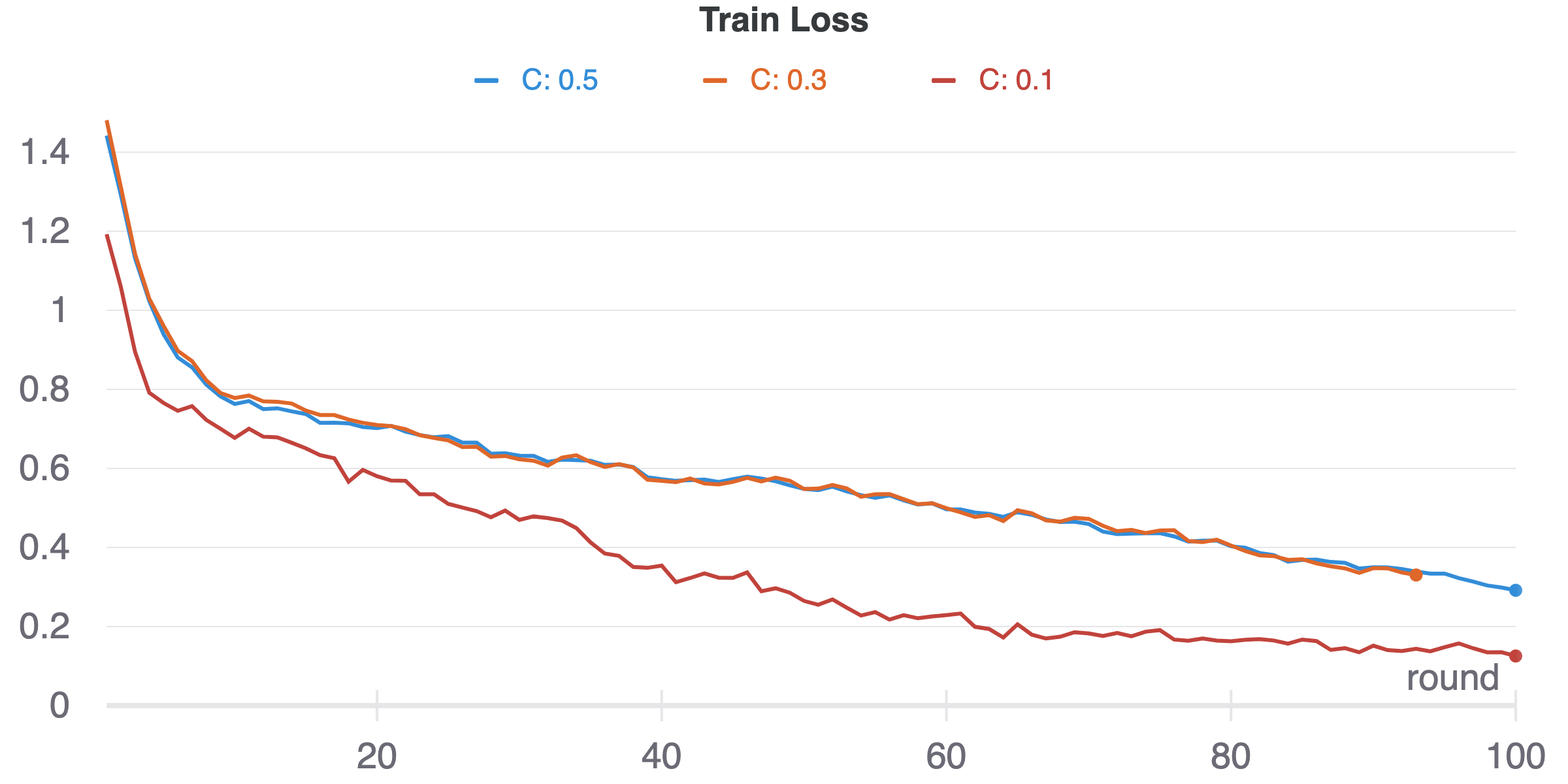}
\end{figure}
\vspace{2pt}

\begin{figure}[H]
\centering
\includegraphics[width=0.8\columnwidth]{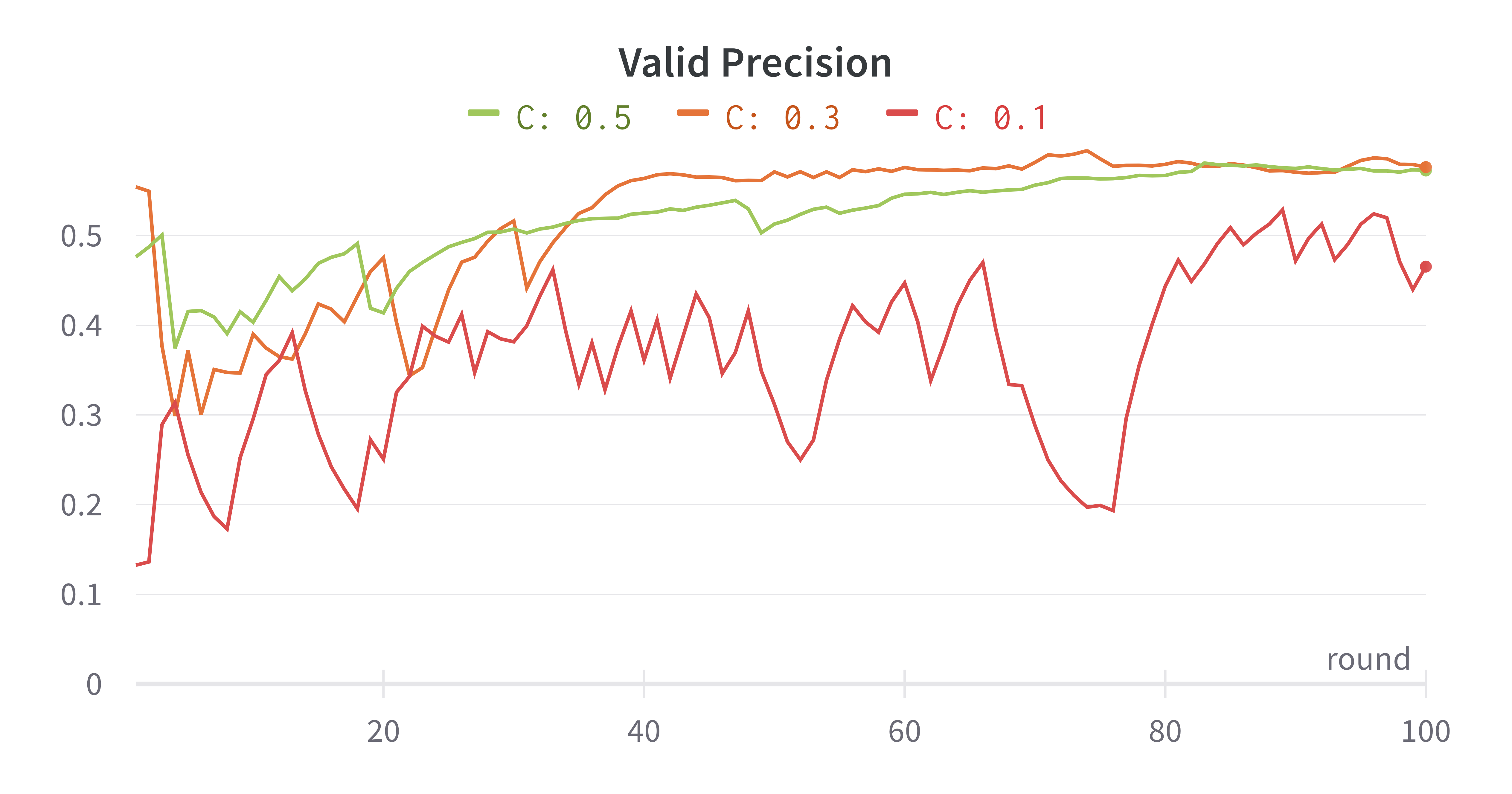}
\end{figure}
\vspace{2pt}

\begin{figure}[H]
\centering
\includegraphics[width=0.8\columnwidth]{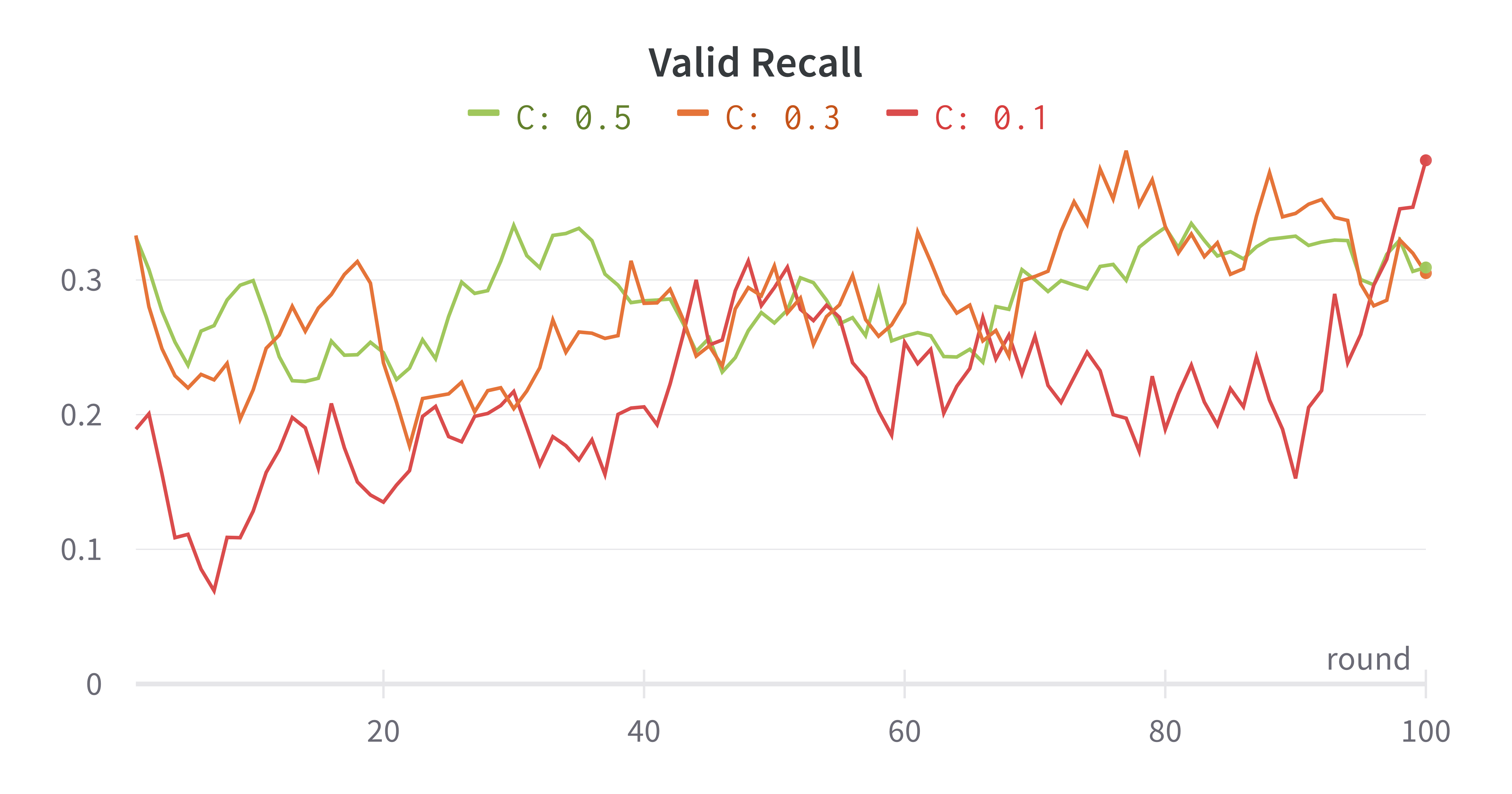}
\end{figure}
\vspace{2pt}

\begin{figure}[H]
\centering
\includegraphics[width=0.8\columnwidth]{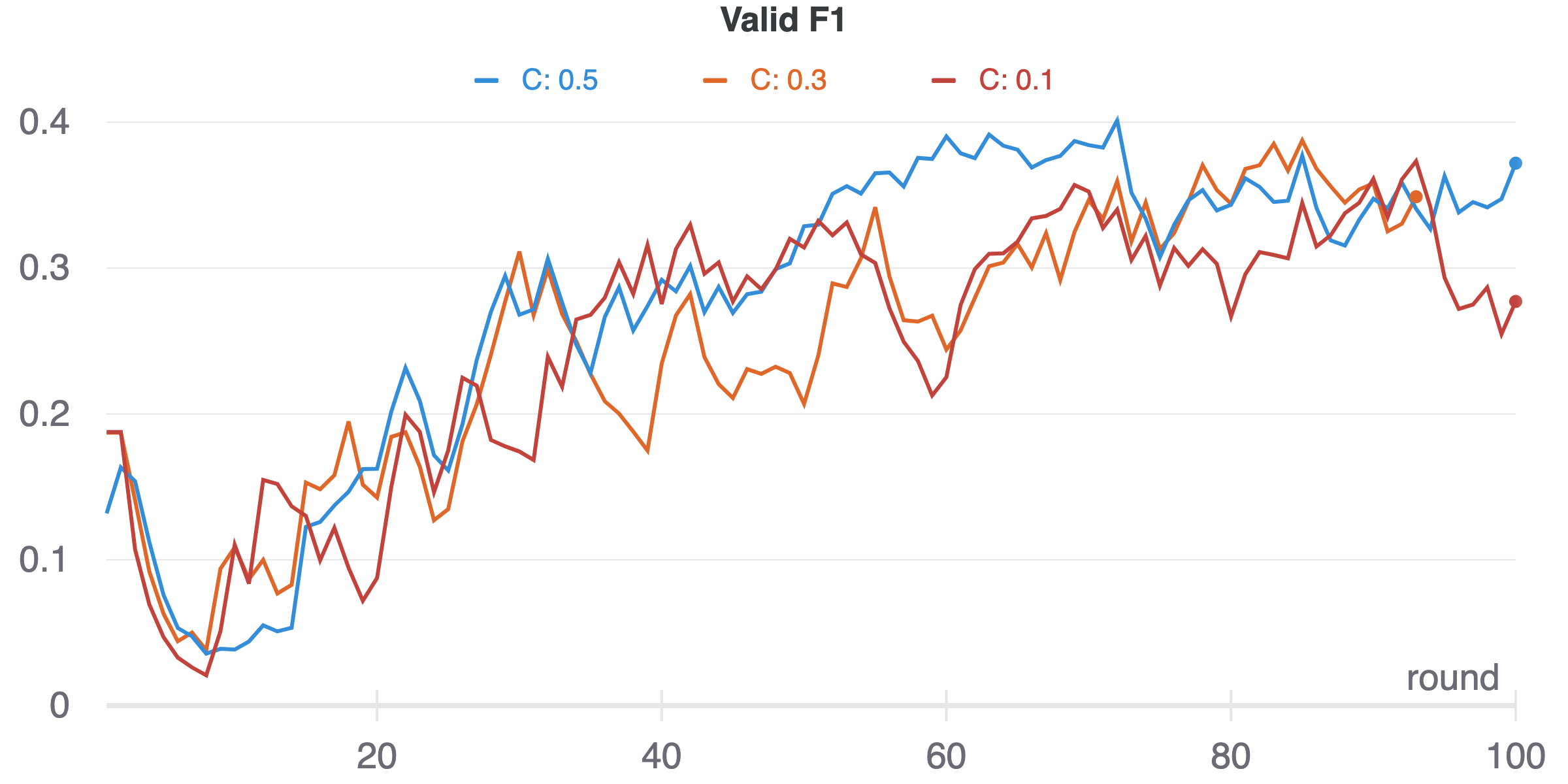}
\end{figure}
\vspace{2pt}

\subsubsection{Cost Sensitive Approach (IID)}

\begin{figure}[H]
\centering
\includegraphics[width=0.8\columnwidth]{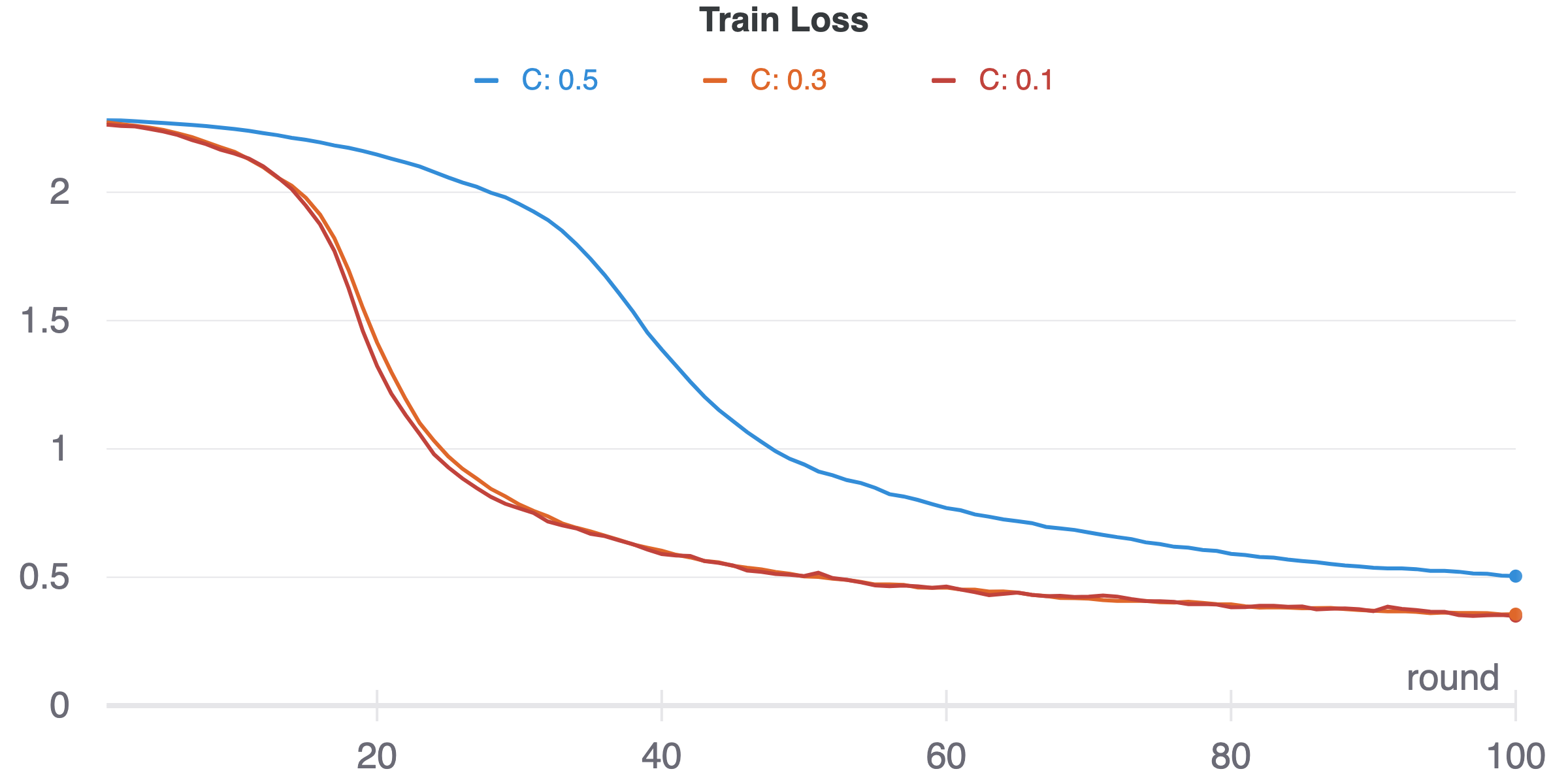}
\end{figure}
\vspace{2pt}

\begin{figure}[H]
\centering
\includegraphics[width=0.8\columnwidth]{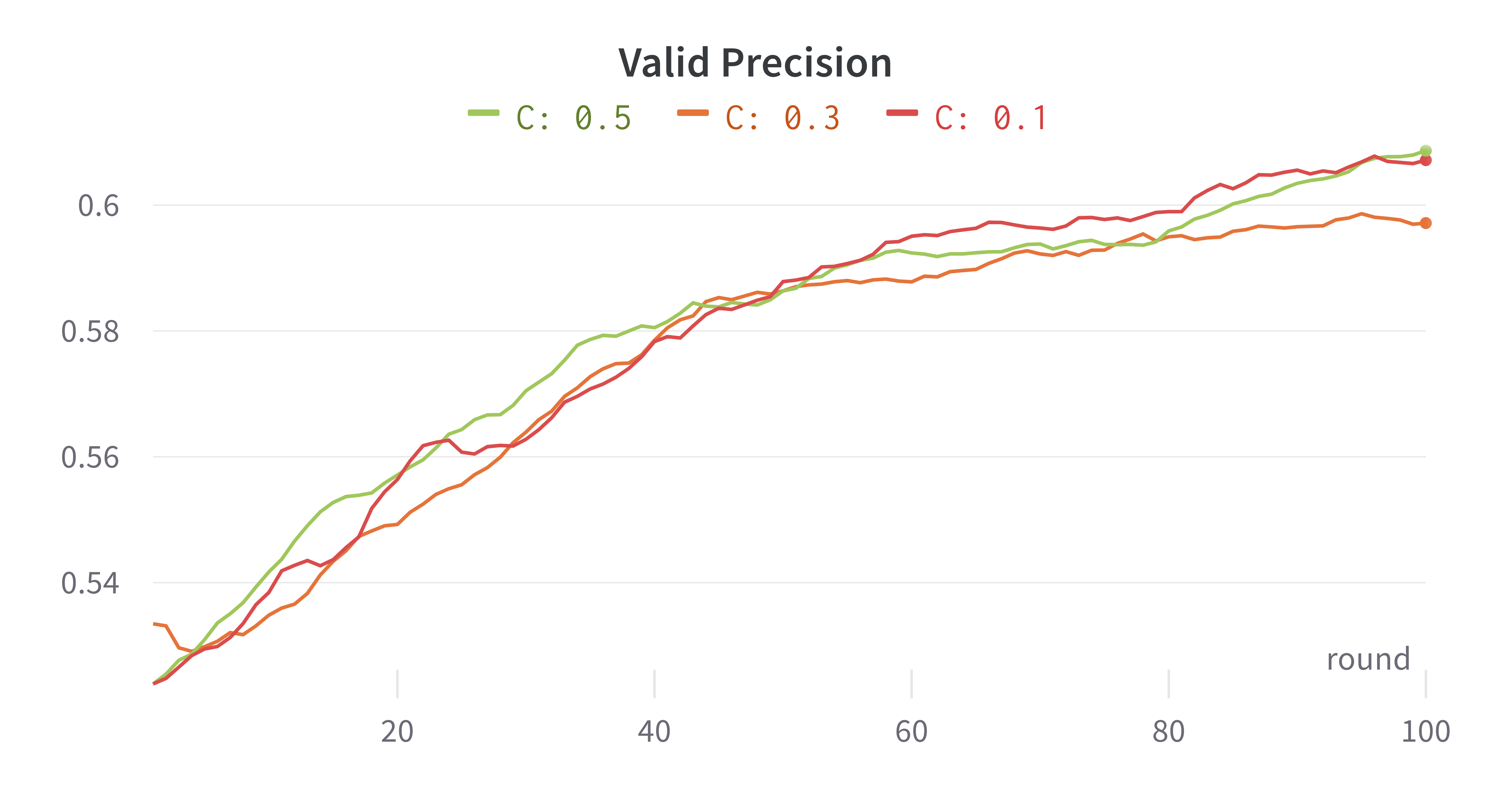}
\end{figure}
\vspace{2pt}

\begin{figure}[H]
\centering
\includegraphics[width=0.8\columnwidth]{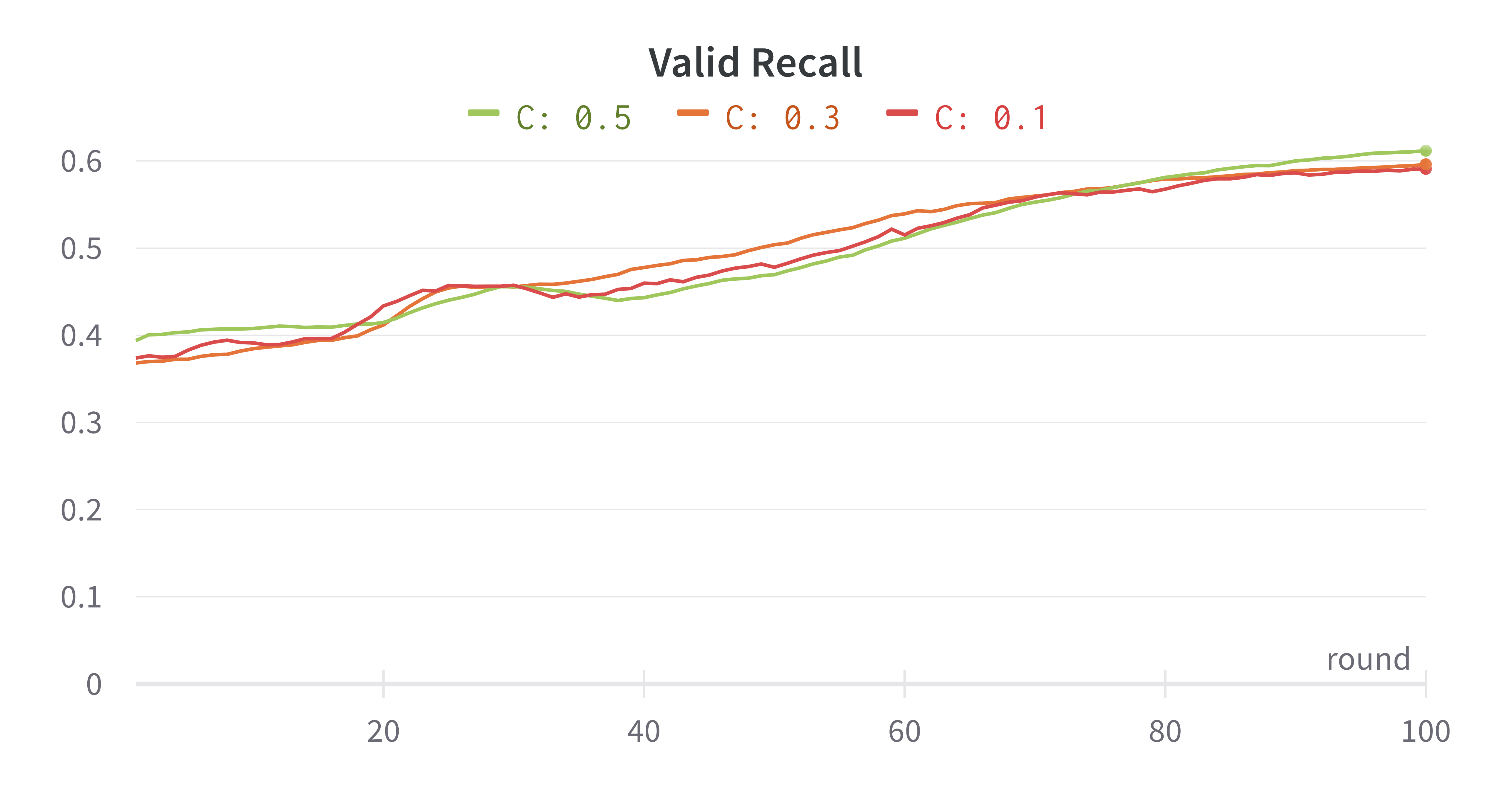}
\end{figure}
\vspace{2pt}

\begin{figure}[H]
\centering
\includegraphics[width=0.8\columnwidth]{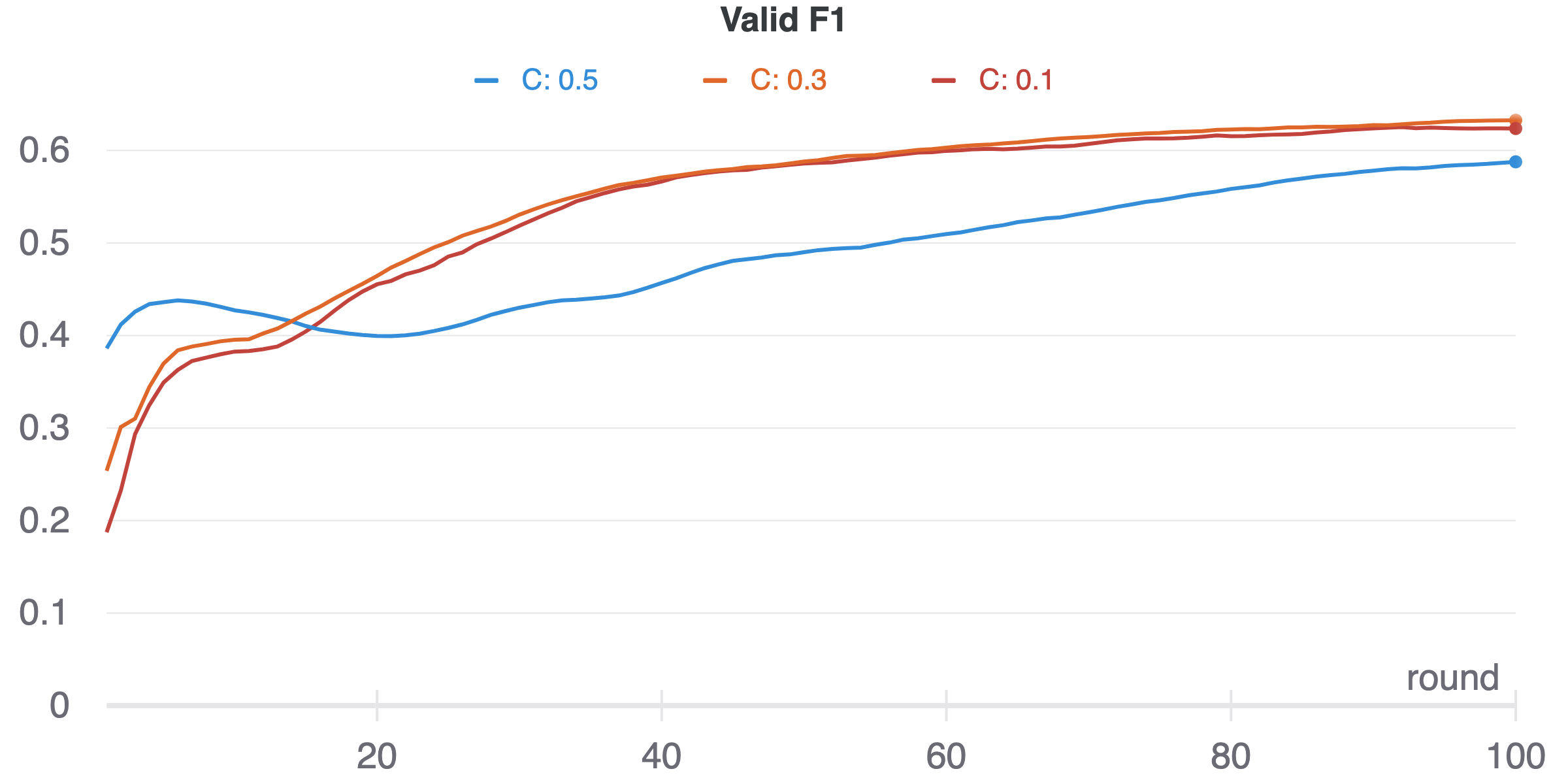}
\end{figure}
\vspace{2pt}

\subsubsection{Cost Sensitive Approach (non-IID)}

\begin{figure}[H]
\centering
\includegraphics[width=0.8\columnwidth]{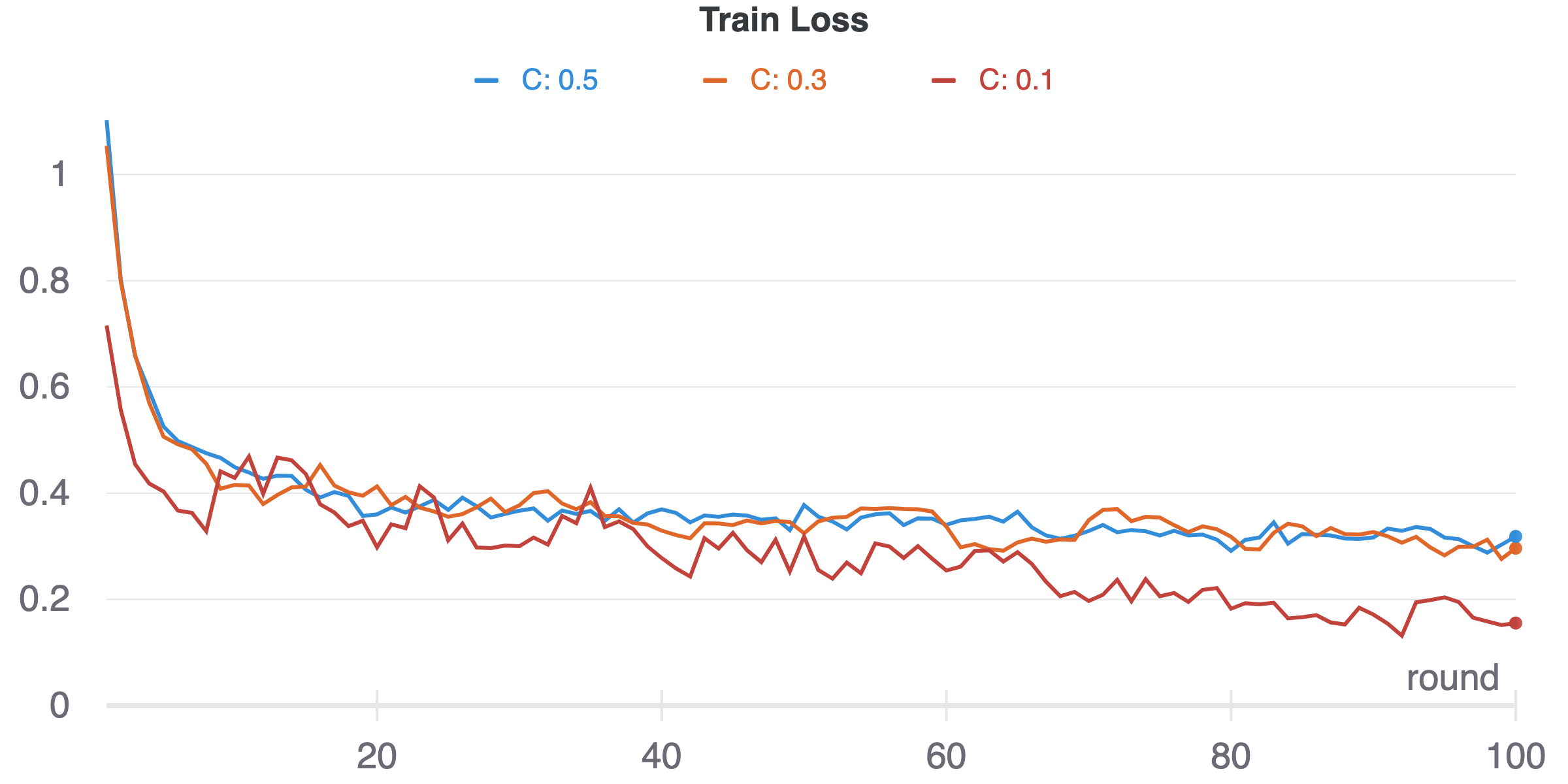}
\end{figure}
\vspace{2pt}

\begin{figure}[H]
\centering
\includegraphics[width=0.8\columnwidth]{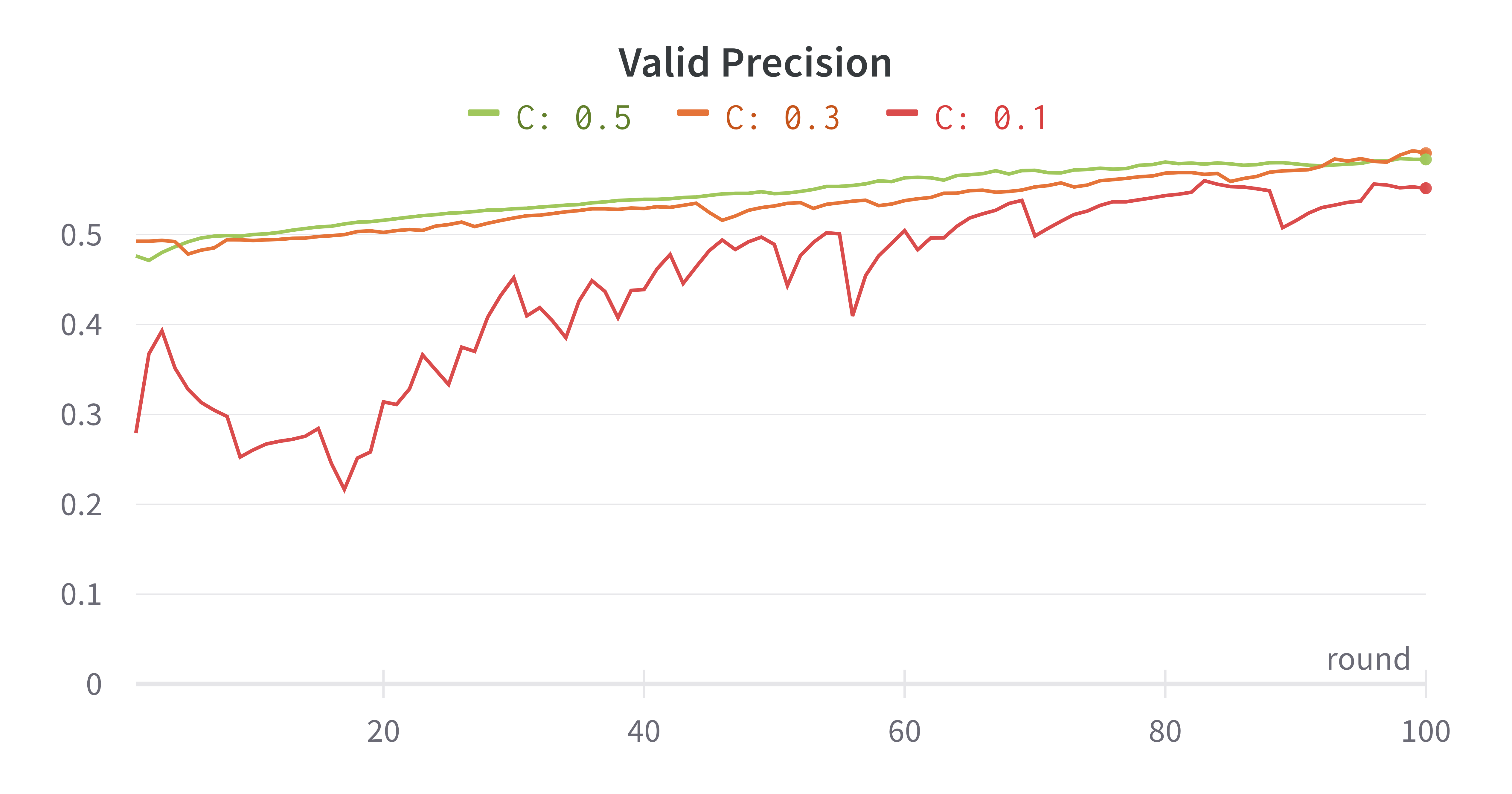}
\end{figure}
\vspace{2pt}

\begin{figure}[H]
\centering
\includegraphics[width=0.8\columnwidth]{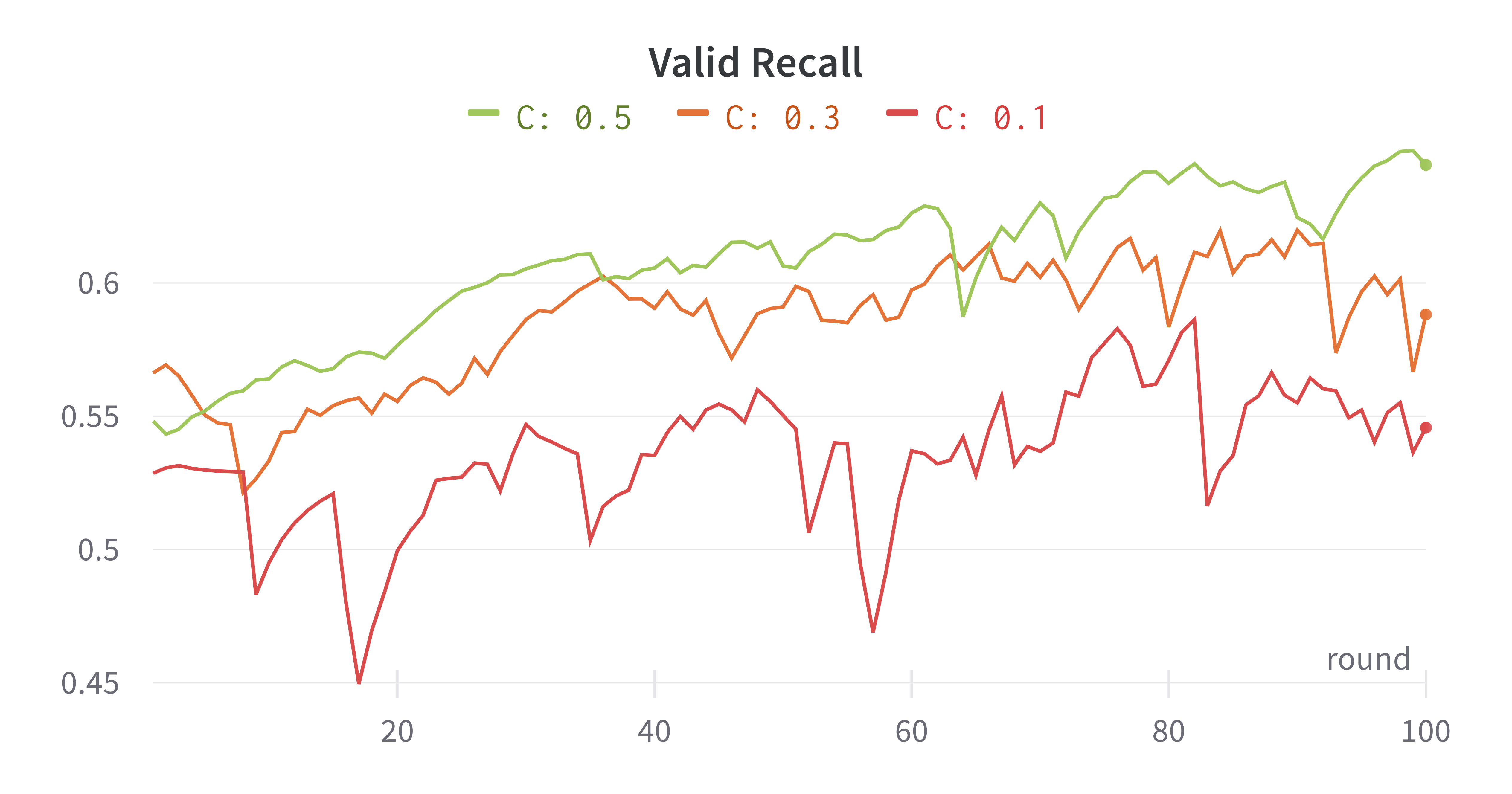}
\end{figure}
\vspace{2pt}

\begin{figure}[H]
\centering
\includegraphics[width=0.8\columnwidth]{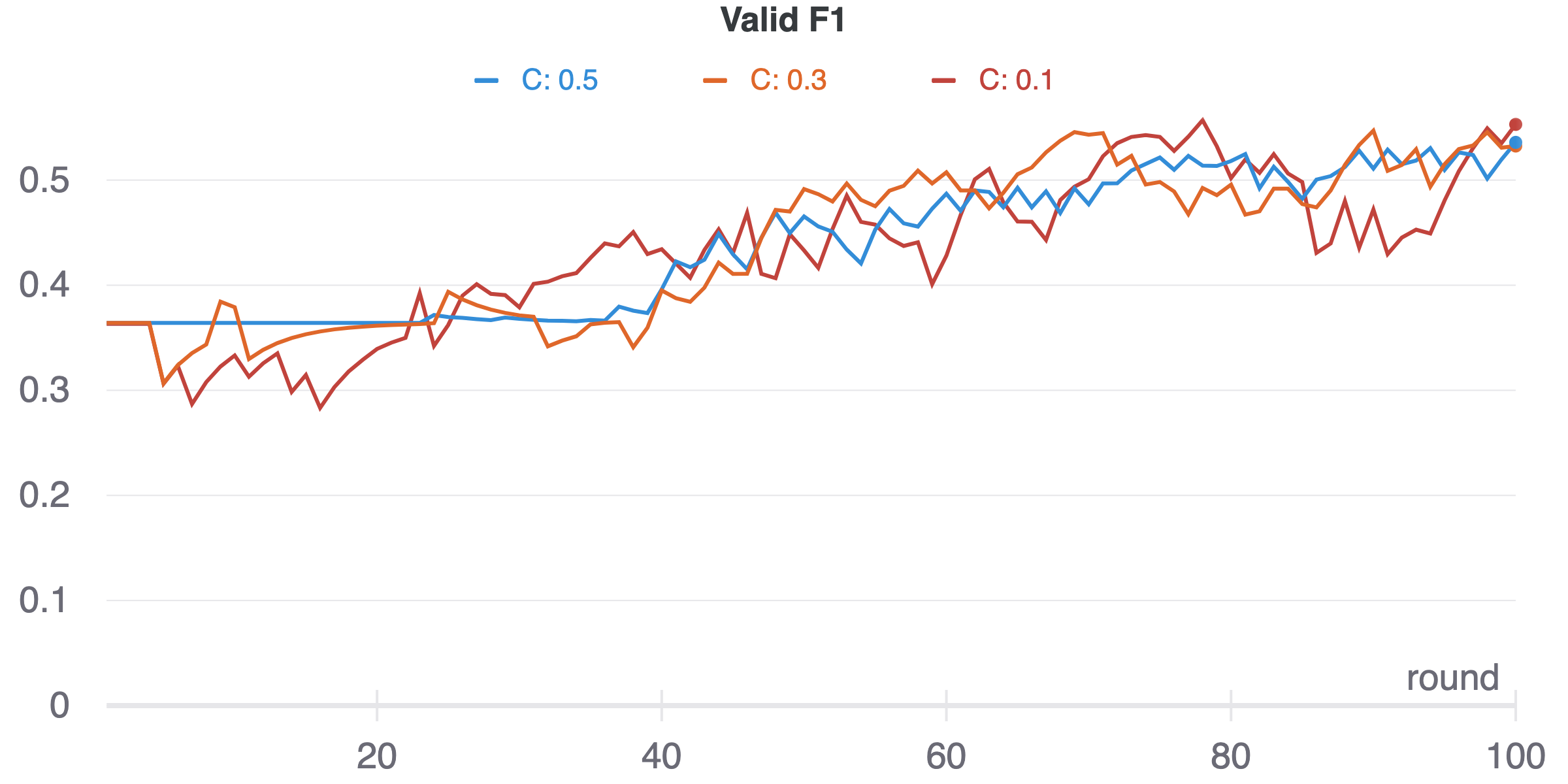}
\end{figure}
\vspace{2pt}

\subsection{Time vs GPU Usage}
This section provides detailed graphs for GPU usage in Watts for every variation of experiments run.

\subsubsection{Imbalanced Dataset}

\begin{figure}[H]
\centering
\includegraphics[width=0.8\columnwidth]{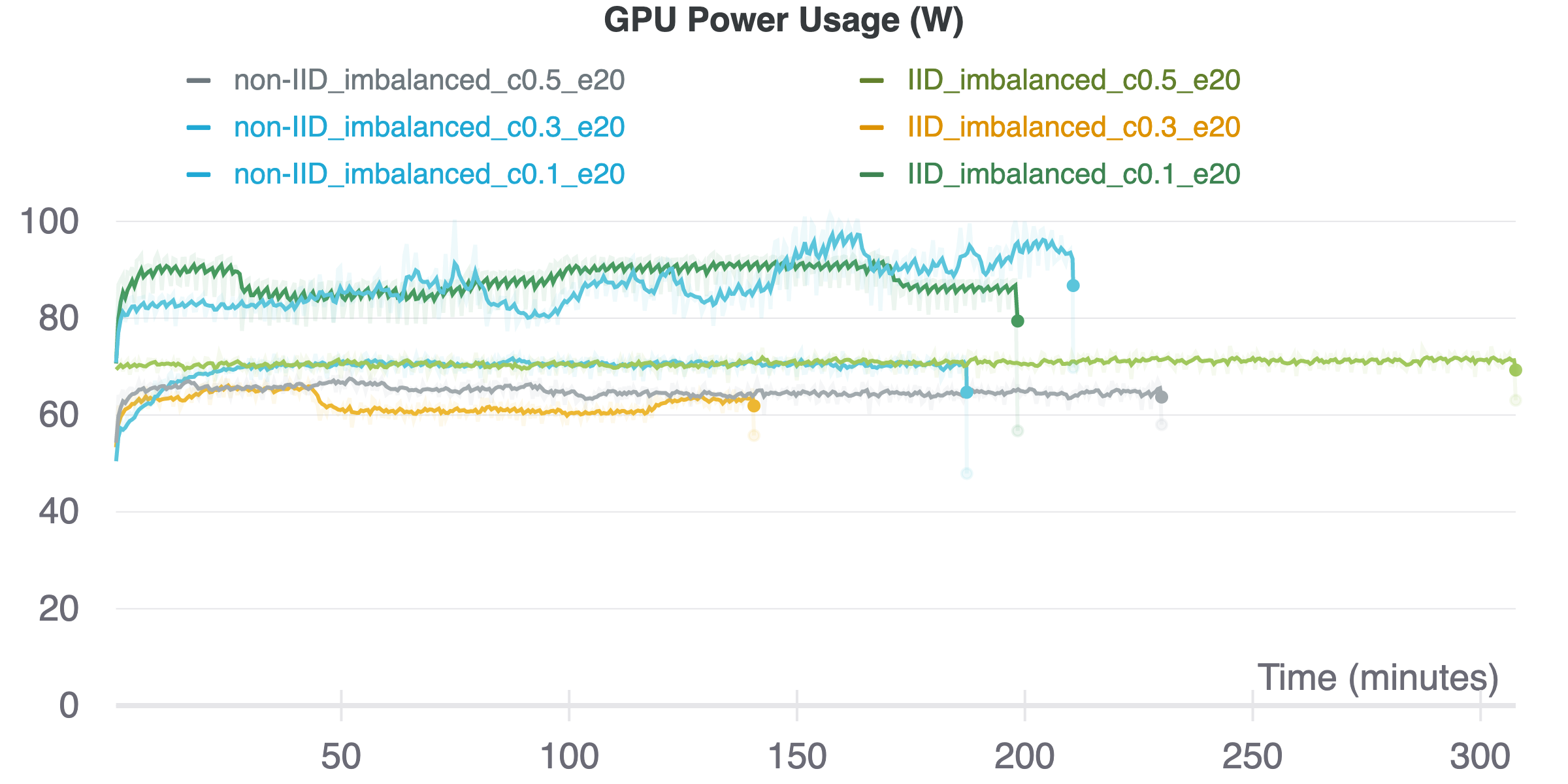}
\end{figure}
\vspace{2pt}

\subsubsection{Balanced Dataset}

\begin{figure}[H]
\centering
\includegraphics[width=0.8\columnwidth]{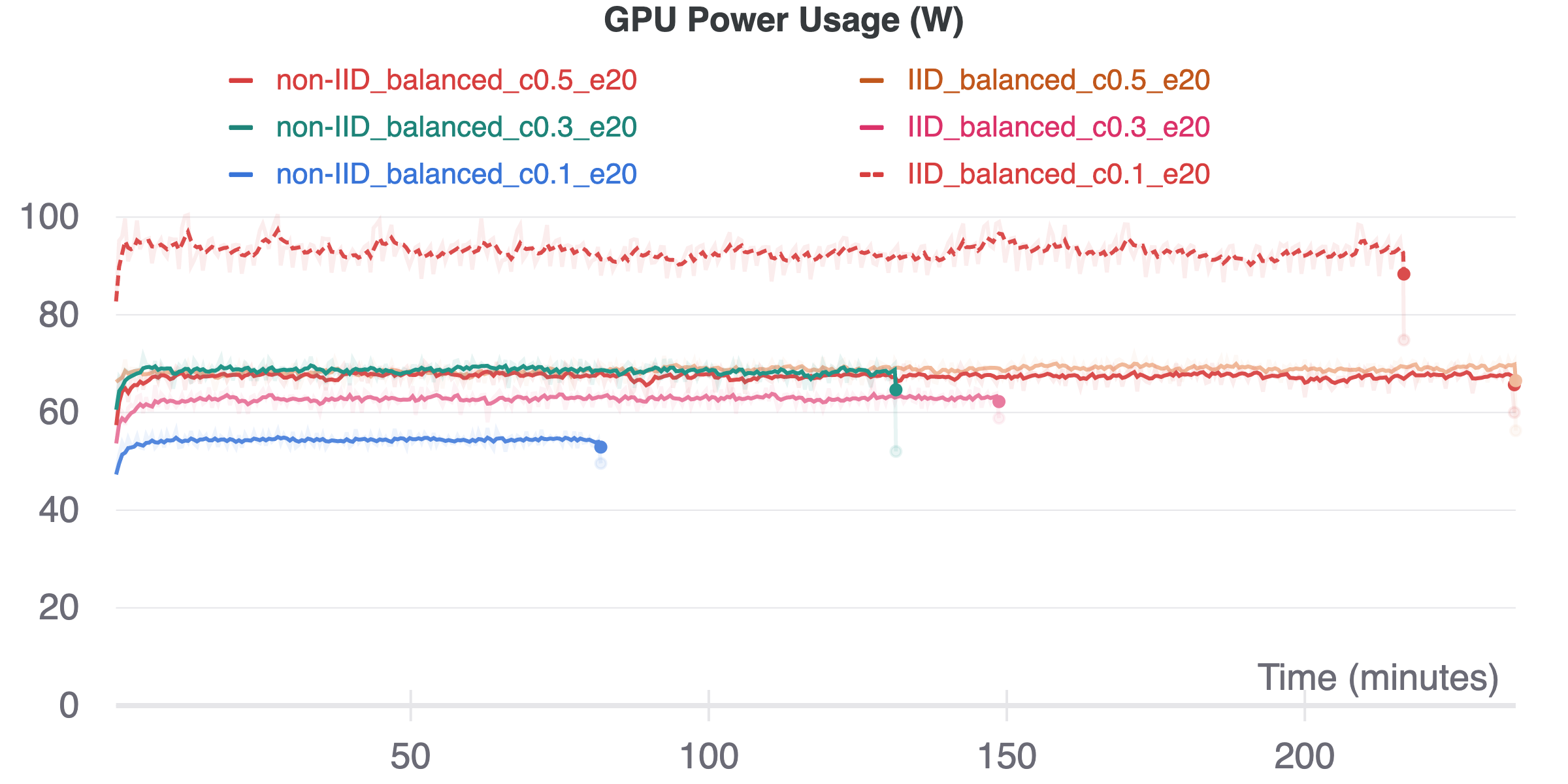}
\end{figure}
\vspace{2pt}

\subsubsection{Cost Sensitive Approach}

\begin{figure}[H]
\centering
\includegraphics[width=0.8\columnwidth]{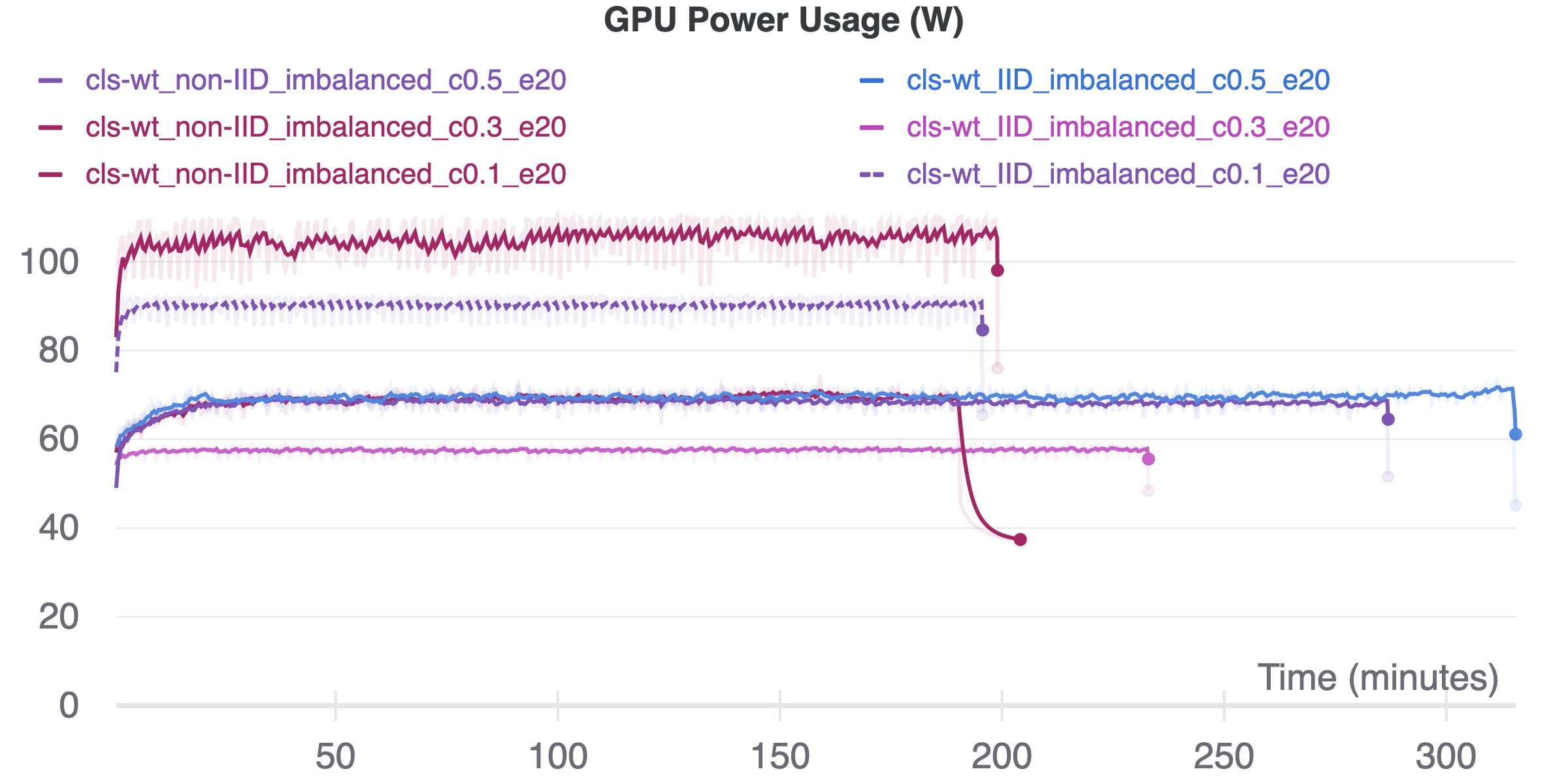}
\end{figure}
\vspace{2pt}

\subsection{Performance Analysis of CausalFedGSD vs Modified CausalFedGSD}
\label{sec:performance_comparison}
\begin{figure}[H]
\centering
\includegraphics[width=0.8\columnwidth]{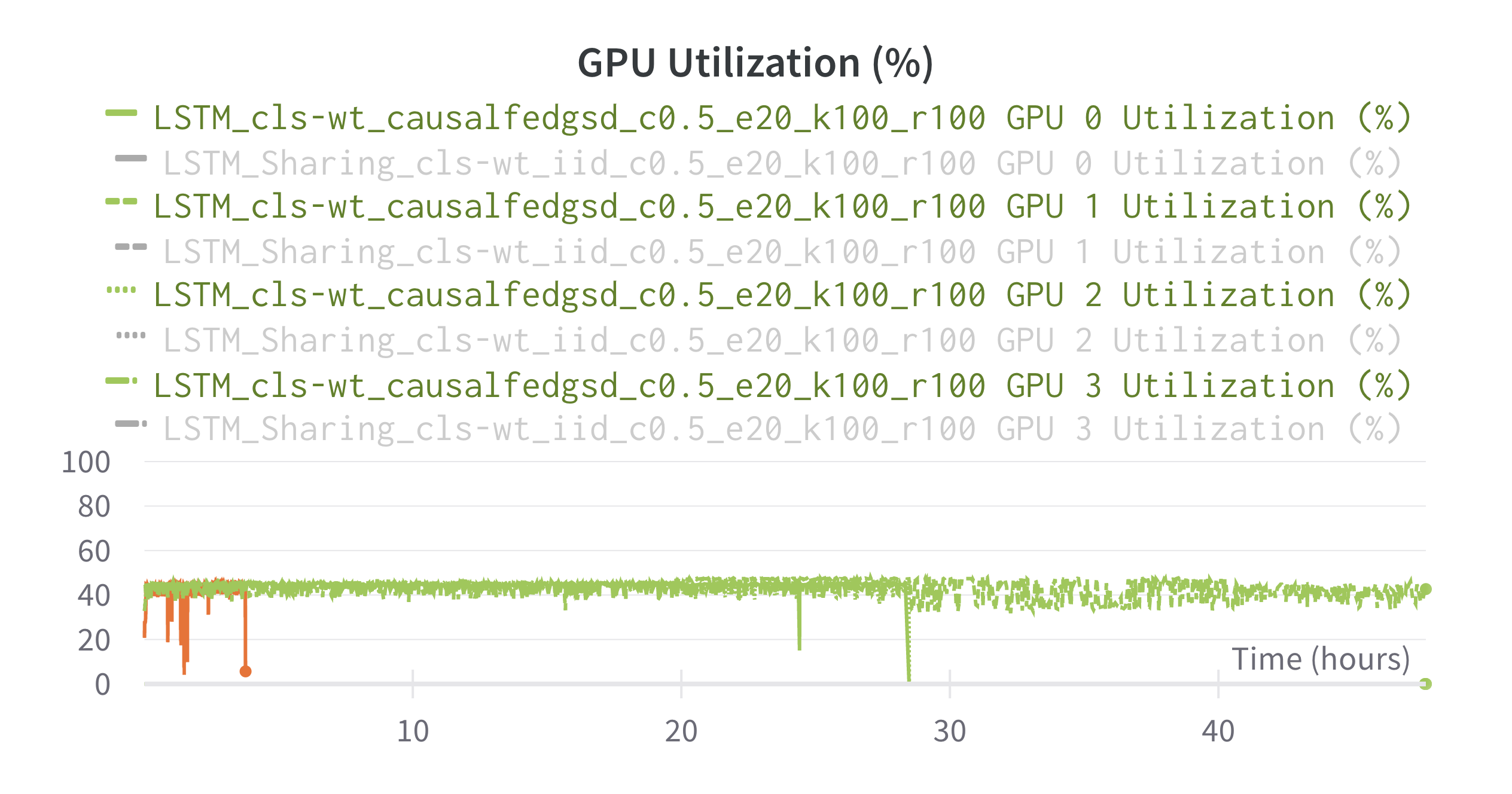}
\caption{Comparison between optimization times for the baseline CausalFedGSD vs Modified CausalFedGSD}
\label{fig:gpu_util}
\end{figure}

We observe that when we run the original CausalFedGSD and our modification on the same hardware settings with the same number of parameters, the modified version finishes training in just 3.75 hours as opposed to the original CausalFedGSD implementation which takes around 47 hours to finish training. Figure~\ref{fig:gpu_util} is a run comparison based on the heaviest variant for both algorithms. 
The highest runtime recorded for both was for the class weight dataset $c=0.5$ variant. 
Similar performances are recorded for all other variants on all datasets.
\vspace{2pt}
\end{document}